%% file: main.tex
\documentclass[times,final,12pt]{elsarticle}

\usepackage{amssymb}
\usepackage{amsmath}

\usepackage{siunitx}
\usepackage{times}
\usepackage{graphicx}
\usepackage{amsmath, amsfonts}
\usepackage{amssymb}
\usepackage{algorithm}
\usepackage{algorithmic}
\usepackage{multirow}
\usepackage{bm}
\usepackage{float}
\usepackage{lettrine}
\usepackage{stfloats}
\usepackage{cite}
\usepackage[utf8]{inputenc} 
\usepackage[T1]{fontenc}    
\usepackage{hyperref}       
\usepackage{url}            
\usepackage{booktabs}       
\usepackage{amsfonts}       
\usepackage{nicefrac}       
\usepackage{microtype}      
\usepackage{xcolor}         
\usepackage{url}
\usepackage{graphicx}
\usepackage{amsthm}
\usepackage{booktabs}
\usepackage{makecell}
\usepackage[normalem]{ulem}
\usepackage{listings}
\usepackage{pifont}
\usepackage{subcaption}
\usepackage{enumitem}
\usepackage{wrapfig}

\usepackage[dvipsnames]{xcolor} 

\newcommand{\added}[1]{{\color{black}#1}}
\input{math_command}

\newcommand{\mytitle}{PathFinder: Advancing Path Loss Prediction for Single-to-Multi-Transmitter Scenario}
\newcommand{\mdl}{PathFinder}
\newcommand{\myurl}{https://emorzz1g.github.io/PathFinder/}

\journal{XXXXX XXXXX}

\begin{document}

\begin{frontmatter}



\title{\mytitle}




\author[label1,label2]{Zhijie Zhong} 
\author[label1,label2]{Zhiwen Yu\corref{cor1}} 
\author[label1]{Pengyu Li} 
\author[label1]{Jianming Lv} 
\author[label1]{C. L. Philip Chen} 
\author[label1]{Min Chen} 


\cortext[cor1]{Corresponding author. Telephone: +86-20-62893506; Fax: +86-20-39380288; Email: zhwyu@scut.edu.cn.}

\affiliation[label1]{ 
    organization={South China University of Technology},
    addressline={}, 
    city={Guangzhou},
    postcode={510650},
    state={Guangdong},
    country={China}
}

\affiliation[label2]{ 
    organization={Pengcheng Laboratory},
    addressline={}, 
    city={Shenzhen},
    postcode={518066},
    state={Guangdong},
    country={China}
}



\begin{abstract}
Radio path loss prediction (RPP) is critical for optimizing 5G networks and enabling IoT, smart city, and similar applications. However, current deep learning-based RPP methods lack proactive environmental modeling, struggle with realistic multi-transmitter scenarios, and generalize poorly under distribution shifts, particularly when training/testing environments differ in building density or transmitter configurations. This paper identifies three key issues: (1) passive environmental modeling that overlooks transmitters and key environmental features; (2) overemphasis on single-transmitter scenarios despite real-world multi-transmitter prevalence; (3) excessive focus on in-distribution performance while neglecting distribution shift challenges. To address these, we propose PathFinder, a novel architecture that actively models buildings and transmitters via disentangled feature encoding and integrates Mask-Guided Low-Rank Attention to independently focus on receiver and building regions. We also introduce a Transmitter-Oriented Mixup strategy for robust training and a new benchmark, single-to-multi-transmitter RPP (S2MT-RPP), tailored to evaluate extrapolation performance (multi-transmitter testing after single-transmitter training). Experimental results show PathFinder outperforms state-of-the-art methods significantly, especially in challenging multi-transmitter scenarios. Our code and project site are available at: \url{\myurl}
\end{abstract}



\end{frontmatter}



\section{Introduction}
\begin{figure}
    \centering
    \includegraphics[width=0.8\linewidth]{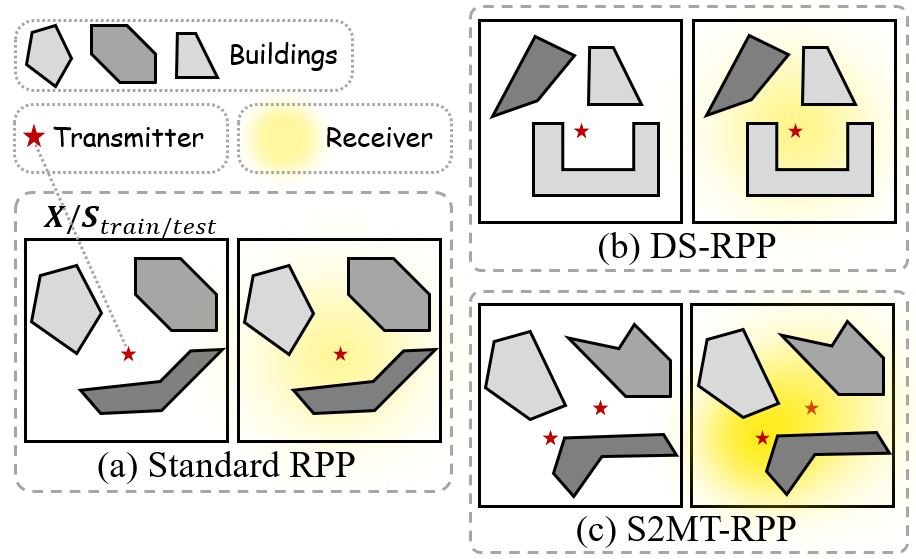}
    \caption{Illustrations of typical RPP task scenarios. \added{(a) Overview of the radio path loss prediction task. (b) Conventional single-transmitter scenario. (c) Proposed single-to-multi-transmitter (S2MT) scenario involving distribution shifts in base station numbers}}
    \label{fig:fw_vs}
\end{figure}

The rapid advancement of 5G communication highlights benefits like low latency and extensive coverage, establishing a fundamental infrastructure for applications in the Internet of Things, intelligent driving, and smart cities \citep{bg_rpp,wifi}. Among these, radio path loss prediction is crucial in communication network planning, signal forecasting, and drone base station/transmitter deployment. In essence, the \textbf{R}adio \textbf{P}ath loss \textbf{P}rediction (RPP) task aims to predict the information loss distribution across regions based on the base station location and relevant data within a specific building environment \citep{rpp2,rpp3,RadioDIP}. This is shown in Fig. \ref{fig:fw_vs}(a).

Challenges in this predictive task often arise from the height and density of buildings in different settings. Additionally, the prediction is complicated by intricate physical phenomena such as diffraction, refraction, and reflection affecting radio signal propagation \citep{RadioUNet,RadioDiff,rpp2}.

The ray tracing method is widely used for predicting wireless point path loss in mobile communication environments \citep{dataset,RadioUNet}. While this method provides accurate predictions, its intricate modeling and computational processes make it unsuitable for real-time applications \citep{RadioUNet,ml_rpp1}. Therefore, it is better suited for design tasks like base station site selection that do not require immediate results. Radio path loss prediction essentially involves estimating or generating predictions.
The increasing need for real-time radio path loss predictions, combined with advancements in deep learning, has sparked significant interest in integrating deep learning methods into this area \citep{ae,dataset,RadioUNet}.
Deep learning-based predictive models \citep{deep_rem,dl_rpp,dl_rpp2} utilize end-to-end learning to reveal implicit relationships between building structures and electromagnetic propagation, resulting in faster and more adaptable model performance for various scenarios. Initially, convolutional neural networks were successfully applied to this task \citep{ae}; however, the lack of extensive validation on real-world datasets was a limitation at the outset, mainly due to the deficiencies in publicly available datasets. Addressing this, \citet{RadioUNet} were the first to tackle the dataset scarcity by introducing the first 2D radio point path loss prediction task and dataset, RadioMapSeer, along with the RadioUNet model for predicting radio path loss. RadioMapSeer, a dataset simulated on a real map using ray tracing software, did not consider building height information or diverse transmitter locations due to algorithm constraints at the time. Recognizing this limitation, subsequent researchers \citet{dataset} developed a new dataset, RadioMapSeer3D, expanding on the RadioMapSeer methodology by incorporating building height information and offering a wider range of transmitter locations. Over time, this dataset has become the primary public dataset in the field \citep{REM-Net,PMNet}.
The development of datasets has led to the emergence of numerous deep learning models \citep{PMNet,REM-Net,unet} for radio path loss prediction, as detailed in the related work section. However, current radio path loss prediction models mainly concentrate on enhancing model performance and refining prediction accuracy, overlooking other crucial aspects of the task. We have pinpointed three vital issues that have been underexplored in previous studies but substantially impede the practical utility of RPP models.

\textbf{Problem 1}: \textbf{Current models demonstrate insufficient proactive modeling of the environment.} Previous studies \citep{RadioUNet,RadioDIP,RadioDiff,PMNet} consider the transmitter as an inherent, static feature on the map. This passive modeling strategy causes the models to disregard the significance of actively incorporating the transmitter, resulting in inadequate learning of the propagation relationship between the transmitter and receiver. 
Additionally, existing research overlooks the independent guidance of the model's focus on building regions and receivers, neglecting their essential roles within the environment. \added{Moreover, while global environmental modeling is crucial for accuracy, traditional attention mechanisms \citep{simad,patchad} impose a prohibitive quadratic computational burden on high-resolution radio maps, limiting their practical deployment.}

\textbf{Problem 2: A substantial gap exists between modeling single-transmitter scenarios and addressing real-world multi-transmitter environments.}
Previous studies \citep{RadioUNet,RadioDIP,RadioDiff,PMNet} have predominantly focused on single-transmitter scenarios (\textit{e.g.}, Fig.~\ref{fig:fw_vs}(b)), while neglecting the prevalence of multiple transmitters in practical settings. However, in real-world 5G deployments, the ``single-to-multi-transmitter (S2MT)'' shift is widespread. For instance, base stations are dynamically added in expanding urban areas. Despite this, existing models lack the ability to extrapolate from single-transmitter training data to multi-transmitter testing scenarios. For example, models trained on data with 1 transmitter often exhibit notable drops in accuracy when tested on scenes with 2–4 transmitters (Fig.~\ref{fig:fw_vs}(c)). This limitation in extrapolation directly restricts the practical utility of current RPP models, rendering it a critical issue that demands resolution.

In order to address the two major issues identified in previous research, we contribute in three aspects:
\begin{enumerate}
    \item To remedy the deficiency in the proactive modeling capabilities of models concerning the environment and address problem 1, we propose \mdl, which utilizes decoupled features to actively model both buildings and transmitters, thereby enhancing the model's attentiveness to these components. 
    \added{Furthermore, we introduce Mask-Guided Low-Rank Attention to tackle the limitations of passive modeling and high computational costs. Specifically, the Mask-Guided mechanism allows the model to independently focus on the distinct physical roles of receiver and building regions through explicit spatial constraints. Meanwhile, the Low-Rank transformation is employed to capture global environmental relationships with linear complexity, ensuring efficient prediction in large-scale urban scenarios.}
    \item To address problem 2, we introduce Transmitter-Oriented Mixup for model training. Based on this approach, we propose a new evaluation benchmark called single-to-multi-transmitter radio path loss prediction (S2MT-RPP). This benchmark enables a more comprehensive assessment of the capabilities of various models.
    \item We conduct a comparative analysis between the proposed \mdl\; and the current state-of-the-art (SOTA) methods, empirically demonstrating that our approach surpasses the performance of existing SOTA techniques.
\end{enumerate}
\begin{figure*}
    \centering
    \includegraphics[width=1\linewidth]{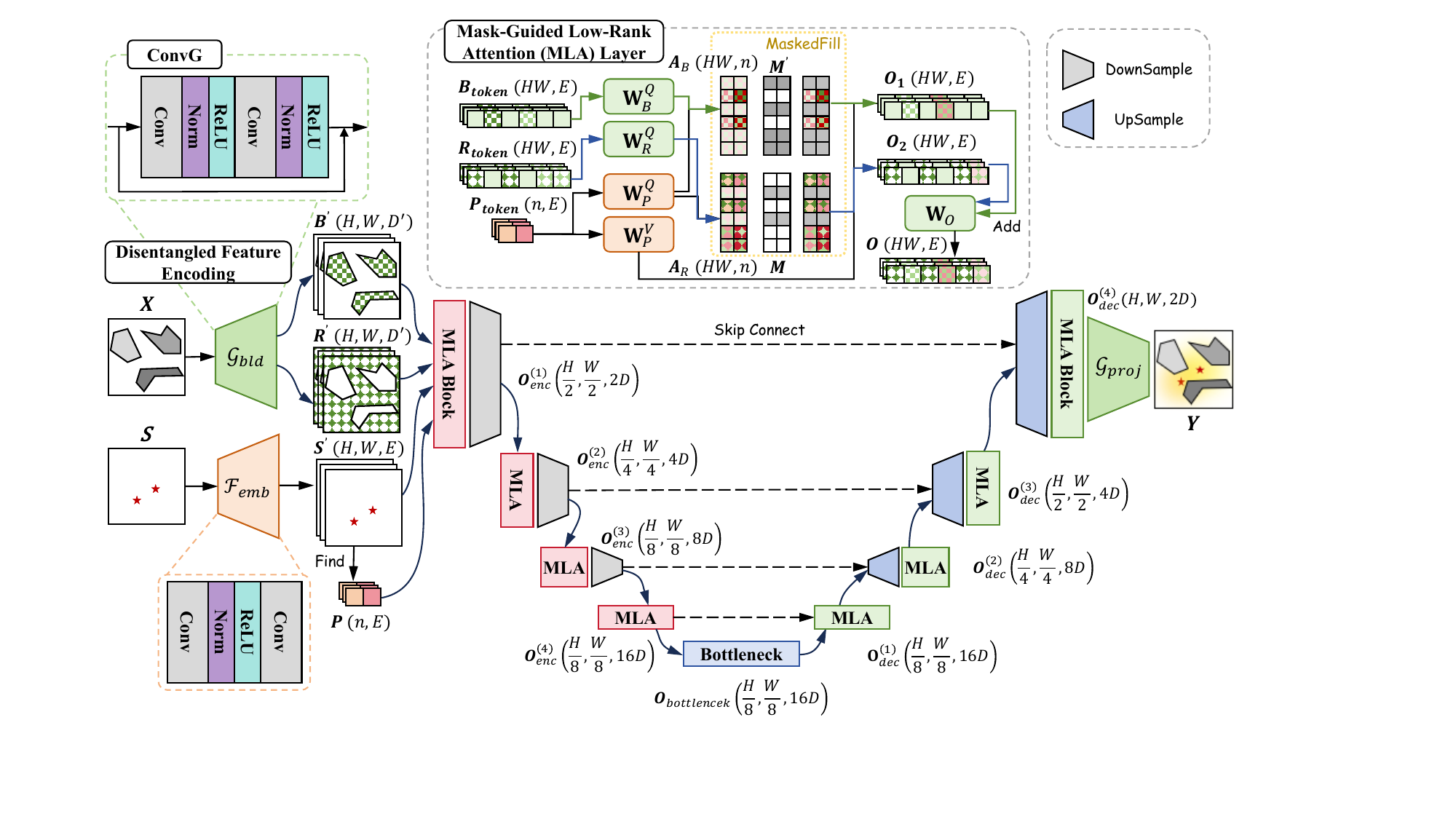}
    \caption{The framework diagram and workflow of PathFinder. \added{The model utilizes Disentangled Feature Encoding (DFE) to decouple building and transmitter features. The UNet structure integrates Mask-Guided Low-Rank Attention (MLA) and Transmitter-Aware Prompts to guide independent focus on building and receiver regions.}}
    \label{fig:fw}
\end{figure*}

\section{Related Work}

\subsection{Radio Pathloss Prediction}
Early deep learning RPP models, limited by dataset scale, saw \citet{ae} construct a convolutional autoencoder (AE) to learn intrinsic data structures through dimensionality reduction \citep{blsg,adamembls}. However, limited parameters restricted the model's expressive capacity, making it suitable only for small scale data scenarios \citep{xi2024umb,Towards_membls}.
Following the RadioMapSeer dataset release, RadioUNet \citep{RadioUNet} first applied UNet to 2D radio path loss prediction, using two independent UNet sub models to learn features from building and transmitter channels. Similarly, DeepREM \citep{deep_rem} used a UNet and conditional generative adversarial network (CGAN) model to estimate RPP from sparse measurements, eliminating the need for additional geographic information. However, this method relies on real radio path loss maps as conditional inputs, limiting its generalization. Additionally, RadioDIP \citep{RadioDIP} predicted path loss in radio signal maps by combining a pre trained RadioUNet with deep image priors (DIP). Nonetheless, this approach did not explore building height's influence on path loss.
With increased demand for prediction accuracy, RadioDiff \citep{RadioDiff} constructed a diffusion model using attention mechanisms and adaptive fast Fourier transforms, significantly enhancing RPP detail representation. However, diffusion models' inherent multi step sampling led to prolonged inference times and increased memory consumption, limiting real time applicability. Moreover, slow inference prevented considering building height information.
Recently, PMNet \citep{PMNet} emerged as the state of the art RPP algorithm, introducing a pyramid structure for multi scale feature extraction and applying dilated convolutions to expand the receptive field. Similarly, REM-Net \citep{REM-Net} achieved precise RPP construction by enlarging the receptive field and mitigating information loss, using multiple sets of dilated convolutions with varying kernel sizes to learn different scale features concurrently. However, these approaches have not incorporated additional information to proactively guide building region modeling, treating them merely as another data processing channel within images.

\subsection{Distribution Shift in Radio Path Loss Prediction}
Machine learning algorithms typically assume training and testing data come from the same distribution. However, real world models often encounter new data that deviates from the training distribution, known as distribution shift. Models trained on a specific distribution may struggle with distribution shift data due to differing features and attributes.

In wireless communication, intricate and variable environments like diverse geographical features, building layouts, and weather conditions cause varying signal propagation. An RPP model trained in one urban area may perform poorly in another city with different terrain or architecture. This typical RPP distribution shift arises from inconsistent environmental or scenario factors between training and testing.

Research on RPP under distribution shift is nascent. \citet{ood_real_time} explored RPP across building regions with distribution shifts, using different architectural maps for training and testing (i.e., differing building layout distributions). \citet{ood_less_is_more} identified another shift: deploying RPP in a region often requires extensive, costly, and challenging sampling. They proposed training models with a subset of RPP samples, as obtaining full samples can be unfeasible. This creates a 'sample availability induced distribution shift' between limited training data and full scene testing.

This study introduces a novel single to multi transmitter distribution shift: models learn from single transmitter RPP maps but must predict cumulative path loss from multiple transmitters during deployment. The core shift lies in the number of transmitters, a key signal propagation factor, between training and testing. To address this, we propose a dataset independent, reusable distribution shift evaluation task: Single to Multi Transmitter Radio Path Loss Prediction (S2MT RPP). This fills a gap in dedicated benchmarks for transmitter related RPP distribution shifts.

\section{Preliminaries}
\subsection{Problem Definition}
Real world propagation involves diffraction, refraction, and reflection. Building materials cause variations in their inherent coefficients. Ray tracing is often too slow for real time radio path loss prediction \citep{dataset}.

Following previous works \citep{PMNet,REM-Net,dataset}, we model actual \(\text{PL}(d)\) as a prediction problem. Given an environmental map \(\mX \in \mathbb{R}^{H\times W}\) with height \(H\) and width \(W\) in meters, \(\mX_{x,y}\) represents the normalized building height, ranging from 0 to 1. The transmitter is a point \((T_x,T_y)\) in the environmental map, translated into an image representation \(\mS \in \mathbb{R}^{H\times W}\). Here, \(\mS_{T_x,T_y}=I\), where \(I\) is the transmitter's height. Model \(f\) learns the relationship between \(\mX\) and \(\mS\) to predict the final radio path loss map \(\mY \in \mathbb{R}^{H \times W}\), expressed as:
\(
    \mY = f(\mX,\mS;\theta),
\)
where \(\theta\) represents the optimizable parameters of model \(f\).

\subsection{Distribution Shift RPP}
In this subsection, we formally define distribution shift radio path loss prediction (DS-RPP) and single-to-multi-transmitter distribution shift radio path loss prediction (MT-RPP), consistent with the distribution shift terminology used in previous sections.

For DS-RPP, we define it as the path loss prediction problem where there is a discrepancy in data distribution between the training and testing phases (the core characteristic of distribution shift). Specifically, during the training phase, the model learns the mapping between the environmental map \(\mX_{\text{train}} \in \mathbb{R}^{H \times W}\) and the transmitter location map \(\mS_{\text{train}} \in \mathbb{R}^{H \times W}\) based on the training data distribution \(\mathcal{D}_{\text{train}}=\{\mX_{\text{train}},\mS_{\text{train}}, \mY_{\text{train}} \}\) to predict the path loss map \(\mY_{\text{train}} \in \mathbb{R}^{H \times W}\). The mathematical expression for this process is:
\(
\mY_{\text{train}} = f(\mX_{\text{train}}, \mS_{\text{train}}; \theta),    
\)
where \(f(\cdot; \theta)\) denotes the model, and \(\theta\) represents the model parameters.

In the testing phase, the model encounters data from a different distribution \(\mathcal{D}_{\text{test}}=\{\mX_{\text{test}},\mS_{\text{test}}, \mY_{\text{test}} \}\) and must predict the path loss map \(\mY_{\text{test}} \in \mathbb{R}^{H \times W}\) based on the testing environmental map \(\mX_{\text{test}} \in \mathbb{R}^{H \times W}\) and the testing transmitter location map \(\mS_{\text{test}} \in \mathbb{R}^{H \times W}\):
\(
\mY_{\text{test}} = f(\mX_{\text{test}}, \mS_{\text{test}}; \theta).    
\)
Here, \(\mathcal{D}_{\text{train}} \neq \mathcal{D}_{\text{test}}\) embodies the core characteristic of distribution shift scenarios.

Furthermore, we define S2MT-RPP as the distribution shift path loss prediction problem in multi-transmitter scenarios. Its mathematical expression is:
\begin{equation}
\mY_{\text{test}} = f\left(\mX_{\text{test}}, \left\{\mS_{\text{test}}^{(1)}, \mS_{\text{test}}^{(2)}, \dots, \mS_{\text{test}}^{(N)}\right\}; \theta\right),   
\end{equation}
where \(\left\{\mS_{\text{test}}^{(1)}, \mS_{\text{test}}^{(2)}, \dots, \mS_{\text{test}}^{(N)}\right\}\) represents the location maps of \(N\) transmitters ( \(N \geq 2\) ), and the key distribution shift lies in the \textbf{number of transmitters} (single in \(\mathcal{D}_{\text{train}}\), multiple in \(\mathcal{D}_{\text{test}}\)).

\section{PathFinder Model}
\mdl consists of three main components: Disentangled Feature Encoding, Transmitter-Aware Prompt, and Mask-Guided Low-Rank Attention. Fig.~\ref{fig:fw} illustrates its architecture and workflow. The environment map \(\mX\) and transmitter \(\mS\) are first processed by Disentangled Feature Encoding (DFE), producing decoupled building features \(\mB^\prime\), transmitter features \(\mS^\prime\), and a Transmitter-Aware Prompt \(\mP\) derived from \(\mS^\prime\). 
At its core, \mdl employs a UNet-based learning network. The encoder includes Mask-Guided Low-Rank Attention (MLA) Blocks and downsample blocks, while the decoder features MLA Blocks and upsample blocks. The bottleneck contains two convolutional layers. In the encoder, each MLA Block (except the last) is followed by a downsample block to halve the feature map size. Similarly, in the decoder, each MLA Block (except the first) is followed by an upsample block to double the feature map size and adjust channel numbers. Finally, the path loss prediction head uses the decoder's final feature maps to estimate path loss.
\subsection{Disentangled Feature Encoding}
During model training, the transmitter's position varies, while building positions remain fixed. Previous methods overlooked this distinction, leading to redundant feature learning \citep{REM-Net,PMNet}. To resolve this, we propose Disentangled Feature Encoding (DFE), which separates building and transmitter feature learning, simplifying feature extraction.

Building maps are represented as a single channel \(\mB \in \mathbb{R}^{H \times W}\), and transmitter information as \(\mS \in \mathbb{R}^{H \times W}\). However, this representation retains redundant transmitter features, complicating dynamic learning. To address this, we introduce a Transmitter-Aware Prompt (Tx-Prompt) method, transforming multiple \(\rs\) data into dynamic prompt vectors.

We first extract transmitter details from \(\mS\), where a single transmitter is \(\rs = (i,j,I)\), with \(i,j\) indicating position and \(I\) the height. Using convolutional feature extraction \(f_{emb}(\cdot)\), \(\mS\) is mapped to high-dimensional features:  
\(
f_{emb}:\mS  \in \mathbb{R}^{H \times W} \rightarrow \mS^\prime \in \mathbb{R}^{H \times W \times E},
\)  
where \(E\) is the embedding dimension. The mapping is defined as:  
\(
\mathcal{F}_{emb}(\mS) = (\operatorname{Conv} \circ \operatorname{BN} \circ \operatorname{ReLU} \circ \operatorname{Conv})(\mS),
\)  
where \(\operatorname{Conv}(\cdot)\) represents 2D convolution, \(\operatorname{BN}(\cdot)\) is batch normalization, and \(\operatorname{ReLU}(\cdot)\) is the activation function.

For \(n\) transmitters \(\sS = \{\rs_1, \rs_2, \cdots, \rs_n\}\), their prompt vectors \(\mP = \{\rp_1, \rp_2, \cdots, \rp_n\}\) are retrieved as \(\mP = \{\operatorname{find}(\mS^\prime,i,j)\}^n\), where \(\operatorname{find}\) indexes the corresponding vector in \(\mS^\prime\). Thus, \(\mP \in \mathbb{R}^{n \times E}\) serves as the Tx-Prompt.

Building features are independently learned using a feature encoder \(g_{bld}\), such that:  
\(
g_{bld}: \mB \in \mathbb{R}^{H\times W} \rightarrow \mB^\prime \in \mathbb{R}^{H \times W \times D^\prime},
\)  
where \(D^\prime = E\). To capture diverse building features, we employ a convolution module \(\operatorname{ConvG}\) with residual connections and group normalization:  
\begin{equation}
\begin{aligned}
    &\mB^\prime = \mathcal{G}_{bld}(\mB) = \operatorname{ConvG}(\mB) + \mB, \\
    &\operatorname{ConvG}(\mB) = (\operatorname{ReLU} \circ \operatorname{GN} \circ \operatorname{Conv} \circ \operatorname{ReLU} \circ \operatorname{GN} \circ \operatorname{Conv})(\mB).
\end{aligned}
\end{equation}
Here, \(\operatorname{GN}(\cdot)\) denotes group normalization.

Ultimately, DFE enables decoupled learning for buildings and transmitters, producing \(\mB^\prime\) and \(\mS^\prime\), along with the transmitter prompt feature \(\mP\).

\subsection{Mask-Guided Low-Rank Attention}
To predict high-frequency information in path loss, we continue with the UNet architecture established in previous works. Specifically, the UNet consists of an encoder, a decoder, and a bottleneck, with both the encoder and decoder comprising \(L(=4)\) Mask-Guided Low-Rank Attention Blocks (MLA Blocks), as illustrated in Fig. \ref{fig:fw}. Each block consists of two \(\operatorname{ConvG}\) layers and one Mask-Guided Low-Rank Attention Layer (MLA Layer). In the feature encoding phase, after each block, a convolutional network is applied to double the channel count of the data while halving the spatial dimensions of the feature maps compared to the previous layer. In the feature decoding phase, the input to the \(j\)-th MLA Block is derived from the output of the \((L-j+1)\)-th layer in the encoder and the output of the \((j-1)\)-th MLA Block. The input to the first MLA Block of the decoder consists of the output from the last layer of the encoder and the output from the bottleneck.
To succinctly describe the workflow of the Mask-Guided Low-Rank Attention Block within each layer, it is important to note that the feature map dimensions \(D\) differ across layers. At the first of each layer, the feature map dimension transitions to \(D^*=2D\), indicating the dimension of the feature map for the subsequent layer, via Eq. (\ref{eq:first_reshape}). 
For a better intuitive understanding of the variant in dimension, refer to Fig. \ref{fig:fw}.

To enable \mdl\; to learn the global information of the environmental region, we combine the features \(\mB^\prime\) and \(\mS^\prime\), resulting in \(\mZ= \mX^\prime(\mB^\prime + \mR^\prime) +\mS^\prime\) as the input to each \(\operatorname{ConvG}\). Thus, the input to the MLA Layer can be expressed as:
\begin{equation}
    \mX_1 = (\operatorname{ConvG} \circ \operatorname{ConvG})(\mZ), (\mathbb{R}^{H\times W \times D^\prime} \rightarrow \mathbb{R}^{H\times W \times D}).
    \label{eq:first_reshape}
\end{equation}
Here, \(D^\prime\) and \(D\) denote the dimensions of the current layer and the previous layer.
For the first layer, \(D^\prime\) is the original dimension of building features.
However, this approach alone does not explicitly guide the model in learning how to predict radio path loss, as it does not provide clear guidance on the relationships between the transmitter and the buildings, as well as between the transmitter and the receiver.

To further address this issue, we consider the environmental map \(\mX_1\) as comprising two components: the building region \(\mB^\prime\) and the non-building region, which corresponds to the receiver region \(\mR^\prime\). By utilizing a mask to guide the model in independently computing attention scores, we introduce Mask-Guided Low-Rank Attention to explicitly learn the global relationships between the building region \(\mB^\prime\) and the transmitter prompts \(\mP\), as well as between the receiver region \(\mR^\prime\) and the transmitter prompts \(\mP\). Unlike traditional self-attention mechanisms, this method allows for more efficient and rapid optimization. The segmentation of the original environmental map significantly reduces the sequence length of the attention, while the Tx-Prompt-guided approach further lowers the computational complexity of cross-attention. Additionally, to enable the model to capture key low-rank information and enhance computational efficiency, we first perform a low-rank transformation \(\mathcal{F}_{LR}\) on the environmental map \(\mX_1\), expressed as:
\begin{equation}
    \mX_{LR} = \mathcal{F}_{LR}(\mX_1) = (\operatorname{Conv} \circ \operatorname{BN})(\mX_1).
\end{equation}

Before constructing the MLA, we unify different features into sequence tokens using reshape and Layer Norm (\(\operatorname{LN}(\cdot)\)) \citep{layerNorm1,layerNorm2}. The building region feature \(\mB^\prime\), the receiver region feature \(\mR^\prime\), and the Tx-Prompt \(\mP\) are converted into building sequence tokens, receiver sequence tokens, and Tx-Prompt tokens, respectively:
\begin{equation}
\small
\begin{aligned}
    &\mB_{token} = (\operatorname{LN}\circ \operatorname{Reshape} \circ \mathcal{F}_{LR})(\mX_1),\ (\mathbb{R}^{H\times W\times D} \rightarrow \mathbb{R}^{HW\times E})\\
    &\mR_{token} = (\operatorname{LN}\circ \operatorname{Reshape} \circ \mathcal{F}_{LR})(\mX_1),\ (\mathbb{R}^{H\times W\times D} \rightarrow \mathbb{R}^{HW\times E})\\
    &\mP_{token} = \operatorname{LN}(\mP),\ (\mathbb{R}^{n\times E} \rightarrow \mathbb{R}^{n\times E}),
\end{aligned}
\end{equation}
where \(E\) denotes a lower-dimensional space, specifically \(E<D\).

Next, we define two sets of multi-head query matrices to query the relevant features of the building sequence tokens and the receiver sequence tokens. The parameters for the former's multi-head query matrix are denoted as \(\tW^{Q}_{B}\), while those for the latter are \(\tW^{Q}_{R}\). The queried features are represented as \(\mQ_{B} = \mB_{token}\tW^{Q}_{B}\) and \(\mQ_{R} = \mR_{token}\tW^{Q}_{R}\). Simultaneously, we define multi-head key and value matrices to represent the relevant features of the Tx-Prompt tokens, denoted as \(\tW_{K_P}\) and \(\tW^V_P\). Thus, the key and value features for the Tx-Prompt tokens are represented as \(\mK_{P} = \mP_{token} \tW^K_P\) and \(\mV_{P} = \mP_{token} \tW^{V}_P\).

Subsequently, we compute the cross-attention \citep{attention_is} scores between the building sequence tokens and the Tx-Prompt tokens, as well as between the receiver sequence tokens and the Tx-Prompt tokens:
\begin{equation}
\begin{aligned}
     \mA_{B} = \frac{\mQ_{B} \mK_{P}^{\top}}{\sqrt{E}},\ 
     \mA_{R} = \frac{\mQ_{R} \mK_{P}^{\top}}{\sqrt{E}}.\\
\end{aligned}
\end{equation}
Assuming the guiding mask for the building region is \(\mM \in \{0,1\}^{HW \times 1}\), the guiding mask for the receiver region is \(\mM^\prime = \mI - \mM\), with both being complementary sets. We can then derive the final cross-attention scores related to the building sequence and receiver tokens:
\begin{equation}
     \mA_{B} = \operatorname{Fill}(\mA_{B},\mM^\prime) \in \mathbb{R}^{HW \times n},
     \mA_{R} = \operatorname{Fill}(\mA_{R},\mM) \in \mathbb{R}^{HW \times n}.
\end{equation}
Here, the \(\operatorname{Fill}\) operation fills the cross-attention scores in the \(\mM^\prime\) region with negative infinity. Consequently, the retained cross-attention scores are solely related to the mask \(\mM\).

Finally, we utilize the cross-attention scores to identify the regions most closely associated with the value features of the building area and the transmitter prompts, as well as the regions of the receiver area that are most correlated with the value features of the transmitter prompts. The final results are then linearly aggregated. This process can be expressed as follows:
\begin{equation}
\begin{aligned}
    & \mO_1 = \operatorname{Softmax}(\mA_{B})\tV_{P},\ (\mathbb{R}^{HW \times n} \rightarrow \mathbb{R}^{HW \times E})\\
    & \mO_2 = \operatorname{Softmax}(\mA_{R})\tV_{P},\ (\mathbb{R}^{HW \times n} \rightarrow \mathbb{R}^{HW \times E}) \\
    & \mO = \operatorname{Conv}\left(\frac{\mO_1+\mO_2}{2}\tW_{O} \right), \ (\mathbb{R}^{HW \times E} \rightarrow \mathbb{R}^{HW \times E}),\\
    & \mO = \operatorname{Reshape}(\mO)\ (\mathbb{R}^{HW \times E} \rightarrow \mathbb{R}^{H\times W \times D}),
\end{aligned}
\end{equation}
where \(\tW_{O}\) is the parameter matrix for linear aggregation. The operation \(\operatorname{Conv}\) serves to restore the features from the low-rank space to the high-dimensional space, while the \(\operatorname{Reshape}\) function converts the features from a token sequence into a feature map.

\subsection{PathFinder Pipeline}
Below is a detailed explanation of how the encoder, bottleneck, and decoder process feature maps, as shown in Fig. \ref{fig:fw}.
In the encoder, except for the last layer, the output feature \(\mO^{(j)}_{enc}\) of the \(j\)-th layer undergoes downsampling:  
\(
\mO^{(j)}_{enc} = \operatorname{DownSample}(\mO^{(j)}_{enc}), \quad (\mathbb{R}^{H \times W \times D} \rightarrow \mathbb{R}^{\frac{H}{2} \times \frac{W}{2} \times D}).
\)
The bottleneck contains two \(\operatorname{ConvG}\) layers, producing the feature:  
\(
\mO_{bottleneck} = (\operatorname{ConvG} \circ \operatorname{ConvG})(\mO^{(L)}).
\)
In the decoder, the input to the \(j\)-th layer, \(\mZ^{(j)}\), is formed by concatenating the previous decoder output \(\mO^{(j-1)}_{dec}\) (or \(\mO_{bottleneck}\) for the first layer) with the corresponding encoder output \(\mO^{(L-j+1)}_{enc}\) along the channel dimension:  
\begin{equation}
\mZ^{(j)}=
\begin{cases} 
[\mO^{(j-1)}_{dec},\mO^{(L-j+1)}_{enc}], & j>1, \\
[\mO_{bottleneck},\mO^{(L-j+1)}_{enc}], & j=1.
\end{cases}
\end{equation}

Except for the first layer, the output \(\mO^{(j)}_{dec}\) of each decoder layer is upsampled, doubling the feature map size:  
\(
\mO^{(j)}_{dec} = \operatorname{UpSample}(\mO^{(j)}_{dec}), \quad (\mathbb{R}^{H \times W \times D} \rightarrow \mathbb{R}^{2H \times 2W \times D}).
\)
Finally, the path loss prediction head \(\mathcal{G}_{pred}\) estimates the radio path loss:  
\(
\hat{\mY} = \mathcal{G}_{pred}(\mO^{(L)}_{dec}).
\)
\subsection{Transmitter-Oriented Mixup}
In the field of computer vision, data augmentation methods \citep{aug1,aug2,aug3} are typically designed to enhance the model's generalization capability by expanding the dataset \citep{aug_improv}. In the context of path loss prediction, common techniques such as image rotation, horizontal, vertical, and diagonal flipping have been employed and proven effective. However, these methods represent generic strategies from the image processing domain and lack task-specific targeting, failing to further enhance the model's path loss prediction abilities. To overcome this limitation, we draw inspiration from Mixup \citep{aug2} and propose the Transmitter-Oriented Mixup (TOM) data augmentation, grounded in the principle of additivity in signal propagation \citep{signalProp1,signalProp2}. Additionally, we establish new complex scenario tests based on this strategy to evaluate the generalization capabilities of different models, with detailed descriptions provided in the experimental section.

To fully leverage the contributions of different transmitters and their generated path losses, let us assume the original environmental map is \(\mX\), which may contain a set of \(n\) transmitters \(\sS = \{\rs_1, \rs_2, \cdots, \rs_n\}\). The corresponding radio path loss maps generated by these transmitters are denoted as \(\sY= \{\mY_1,\mY_2,\cdots, \mY_n\}, \mY_i\in \mathbb{R}^{H \times W}\). Given any two transmitters \(i\) and \(j\) along with their radio path loss maps \(\mY_i\) and \(\mY_j\), according to the principle of additivity in signal propagation, if the weights of the transmitters are \(a\) and \(b\), the resulting combined radio path loss map can be expressed as \(\mY=a\mY_i+b\mY_j\). This implies that, under the same environmental map conditions, the mixed radio path loss map is solely related to the transmitters. Based on this principle, we propose the Transmitter-Oriented Mixup data augmentation.

Specifically, given any base image augmentation function \(f_{aug}\) (including image rotation, horizontal, vertical, and diagonal flipping, \textit{etc.}), and sampling mix weights \(\beta\sim \operatorname{Beta}(\alpha,\alpha),\beta\in [0,1]\), where \(\beta\) follows a Beta distribution with parameters both set to \(\alpha\), the mixed environmental map can be expressed as
\(\mX_{mixup} = f_{aug}(\mX)\).

Given the set of transmitters \(\sS\), which can be transformed into transmitter maps \(\{\mS_1,\mS_2, \cdots, \mS_n\},\mS_i\in \mathbb{R}^{H\times W}\), we randomly select two transmitter maps \(\mS_i\) and \(\mS_j\). The final mixed transmitter map and radio path loss map are then given by:
\begin{equation}
\begin{aligned}
    \mS_{mixup} = \beta f_{aug}(\mS_i) + (1-\beta) f_{aug}(\mS_j),\\
    \mY_{mixup} = \beta f_{aug}(\mY_i) + (1-\beta) f_{aug}(\mY_j).
\end{aligned}
\label{eq:mixup}
\end{equation}
Finally, we utilize these mixed samples for model training to enhance the model's generalization capability.
\subsection{Model Optimization}
To ensure the accuracy of radio path loss map predictions, the mean squared error (MSE) loss is commonly employed to optimize the model. However, during the early stages of model training, when the model's output significantly deviates from the target, the gradients of the MSE can become excessively large, leading to instability in training. Moreover, due to the substantial number of parameters in the UNet architecture, the slow optimization speed in the initial phase is exacerbated by the MSE loss. To address these shortcomings, we propose the Momentum Prediction Loss (MPL). This loss function introduces an additional parameter \(\delta\) for adaptive momentum updates, smoothing the weights of the MSE loss and the mean absolute error (MAE) loss. It can be expressed as follows:
\begin{equation}
    \mathcal{L}=\left\{
    \begin{aligned}
        &\frac{(\mY-\hat{\mY})^2}{2}, &|\mY-\hat{\mY}|\le\delta\\
        &\delta|\mY-\hat{\mY}|-\frac{\delta^2}{2}, &|\mY-\hat{\mY}|>\delta
    \end{aligned}
    \right.
\end{equation}
where \(\hat{\mY}\) represents the model's predicted radio path loss map. 

After each iteration, the parameter \(\delta\) is updated based on the average predicted MAE loss momentum from the current batch of data (e.g., \(N\) samples) as follows: \(\delta=\max\{0.9\cdot\delta, \frac{\sum_{i}^N|\mY_i-\hat{\mY}_i|}{N}\}\).

\section{Experiment}
\subsection{Experimental Setup}
To facilitate comparison with SOTA methods, we utilize the publicly available RadioMap3DSeer (RM3D) dataset \citep{dataset}. RM3D accounts for varying building heights and transmitters deployed at different elevations, with these transmitters positioned at higher floors of the buildings. The dataset is generated through a combination of Intelligent Ray Tracing (IRT) and real-world maps, making it closely aligned with real-world applications \citep{dataset}. After processing, the dataset comprises a total of 700 urban building maps, with each image representing an area of \SI{256}{m} \(\times\) \SI{256}{m}. All simulations are conducted at a minimum resolution of one meter, resulting in final images with dimensions of 256 \(\times\) 256 pixels. Additionally, each urban map considers 80 transmitter locations, culminating in a total of 56,000 simulations.
\added{To evaluate the generalization performance of various models in entirely unseen scenarios, we incorporate the Rural dataset \ref{PMNet}. Given its limited size of only 328 images, this dataset is more suitable for cross-dataset testing rather than model training. This allows us to assess whether models can generalize to building layouts that differ significantly from the urban environments in RM3D.}

\added{To improve generalization, all models are trained with random horizontal and vertical flipping (each with a probability of 0.5) as a default augmentation strategy.}

\subsubsection{Baseline}
AE \citep{ae}, a convolutional autoencoder, predicts radio path loss. Its limited expressive capacity restricts its application to small datasets.

RadioUNet \citep{RadioUNet} was the first 2D radio path loss prediction algorithm. It models building and transmitter features separately, reducing UNet's training complexity. Easily adaptable to new 3D scenarios, it serves as an evaluation baseline.

PMNet \citep{PMNet}, the current state of the art RPP algorithm, uses a pyramid structure for multi scale feature extraction and dilated convolutions to expand the model's receptive field.

Similarly, REM-Net \citep{REM-Net} achieves precise RPP construction by expanding the receptive field and mitigating information loss. It employs multiple groups of dilated convolutions with varying kernel sizes to learn features at different scales.
\subsubsection{Evaluation Metric}
To evaluate the performance of different methods, we construct a total of six metrics. The first three metrics are global coverage metrics, namely MSE, RMSE, and NMSE. The latter three metrics focus on the receiver area and are referred to as MSE-R, RMSE-R, and NMSE-R.
Due to space limitations, their calculation methods are provided in the appendix.
\subsection{Comparison on DS-RPP}
To compare model performance in DS-RPP scenarios, we partition the RadioMap3DSeer dataset into training, validation, and testing sets at a 5:1:1 ratio, based on sample order. This ensures distinct training and testing distributions.

Table \ref{tab:comparison} shows various models' performance on the testing set. Our model (\mdl) outperforms others across both metrics, achieving SOTA results. It improves RMSE and RMSE-R metrics by 30.59\% and 29.08\% respectively, compared to REM-Net, significantly enhancing prediction performance in both global and local areas.

\begin{figure*}
    \centering
    \includegraphics[width=.9\linewidth]{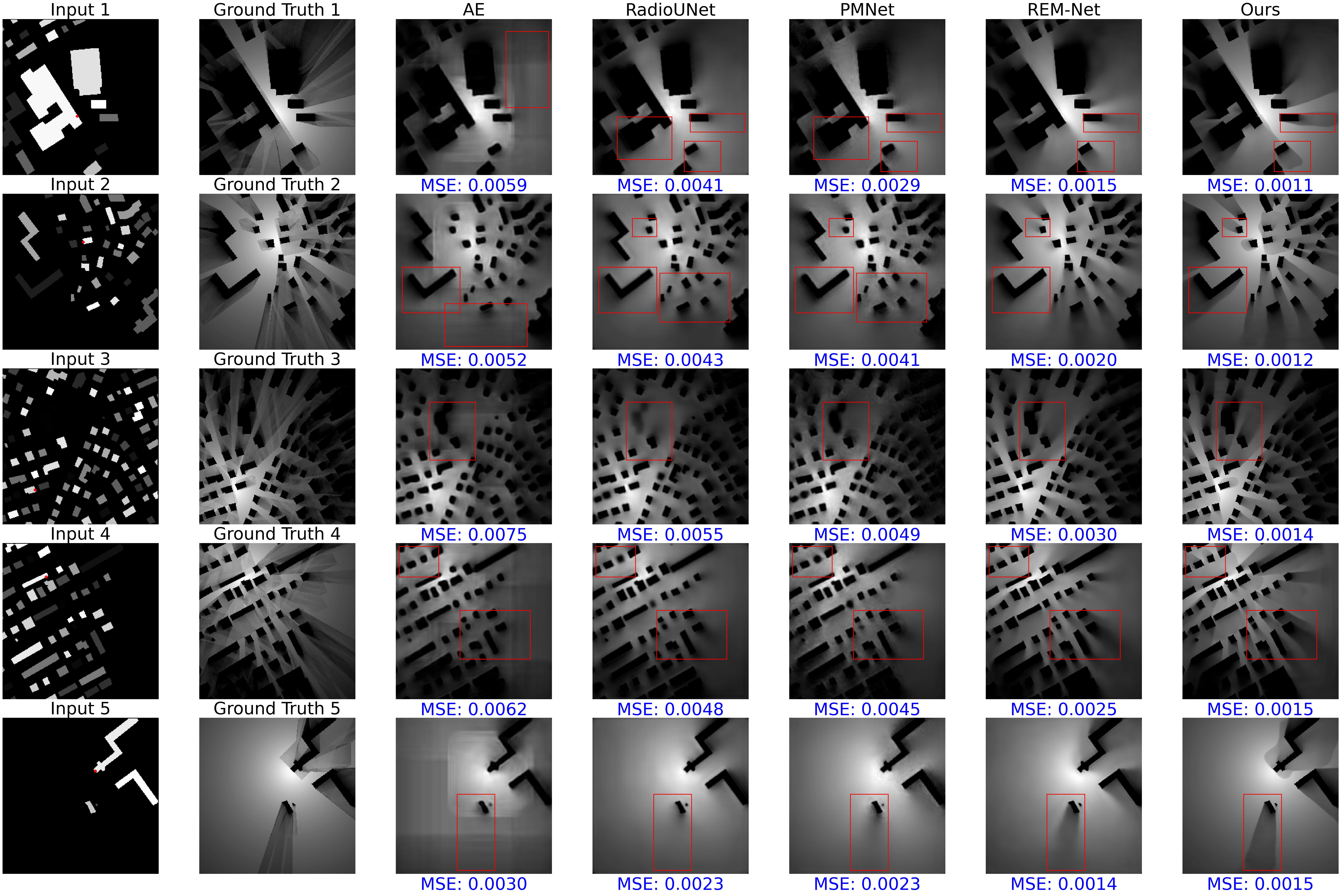}
    \caption{\added{Visual comparison of different models in the DS-RPP task. From left to right: input maps (red dots indicate Tx), Ground Truth, and predictions from various models. Red boxes highlight significant performance gaps. While competing methods show less distinct boundaries and less accurate predictions in complex building layouts, PathFinder demonstrates superior accuracy in boundary definition and complex building areas.}}
    \label{fig:comp_trail1}
\end{figure*}

\added{To intuitively evaluate the predictive performance, Fig. \ref{fig:comp_trail1} presents a visual comparison between PathFinder and various baseline methods across diverse scenarios. As highlighted by the red boxes, while competing methods often produce blurred boundaries or less accurate predictions in complex building layouts, PathFinder consistently achieves superior accuracy in boundary definition. This improvement is particularly evident in densely built areas, where our model effectively captures intricate propagation patterns that others overlook. A more detailed analysis of these performance gaps follows below.}

The AE model's limited receptive field leads to weak predictive capability for distant regions, as seen in Inputs 1 and 2. It also primarily focuses on areas near the transmitter (Input 5), making it the poorest performer among all models.

RadioUNet and PMNet show similar performance with minimal MSE differences. Both struggle with path loss predictions along building edges (Inputs 3, 4, and 5), where predictions around contours appear vague, lacking clear boundaries. PMNet's more complex architecture makes it more prone to overfitting, leading to blurrier predictions in distribution shift scenarios than RadioUNet. Thus, despite PMNet's lower MSE loss, its distribution shift handling ability is comparable to RadioUNet's.

REM-Net is the best-performing baseline, successfully establishing basic path loss boundaries in all samples. However, its boundary delineation lacks the precision of our proposed model. For instance, in Inputs 1 and 5, path loss regions farther from the transmitter show suboptimal clarity, while our model exhibits more distinct boundaries. This improvement stems from the attention mechanism's ability to learn global features, allowing the model to focus on features from distant locations. Additionally, in Input 2, REM-Net's path loss estimation on the building side away from the transmitter is insufficiently accurate. In complex building distribution areas, our model surpasses REM-Net, as seen in the red boxes of Inputs 3 and 4, where multiple buildings are present and our model's path loss boundaries are clearer.

\begin{figure}
    \centering
    \includegraphics[width=1\linewidth]{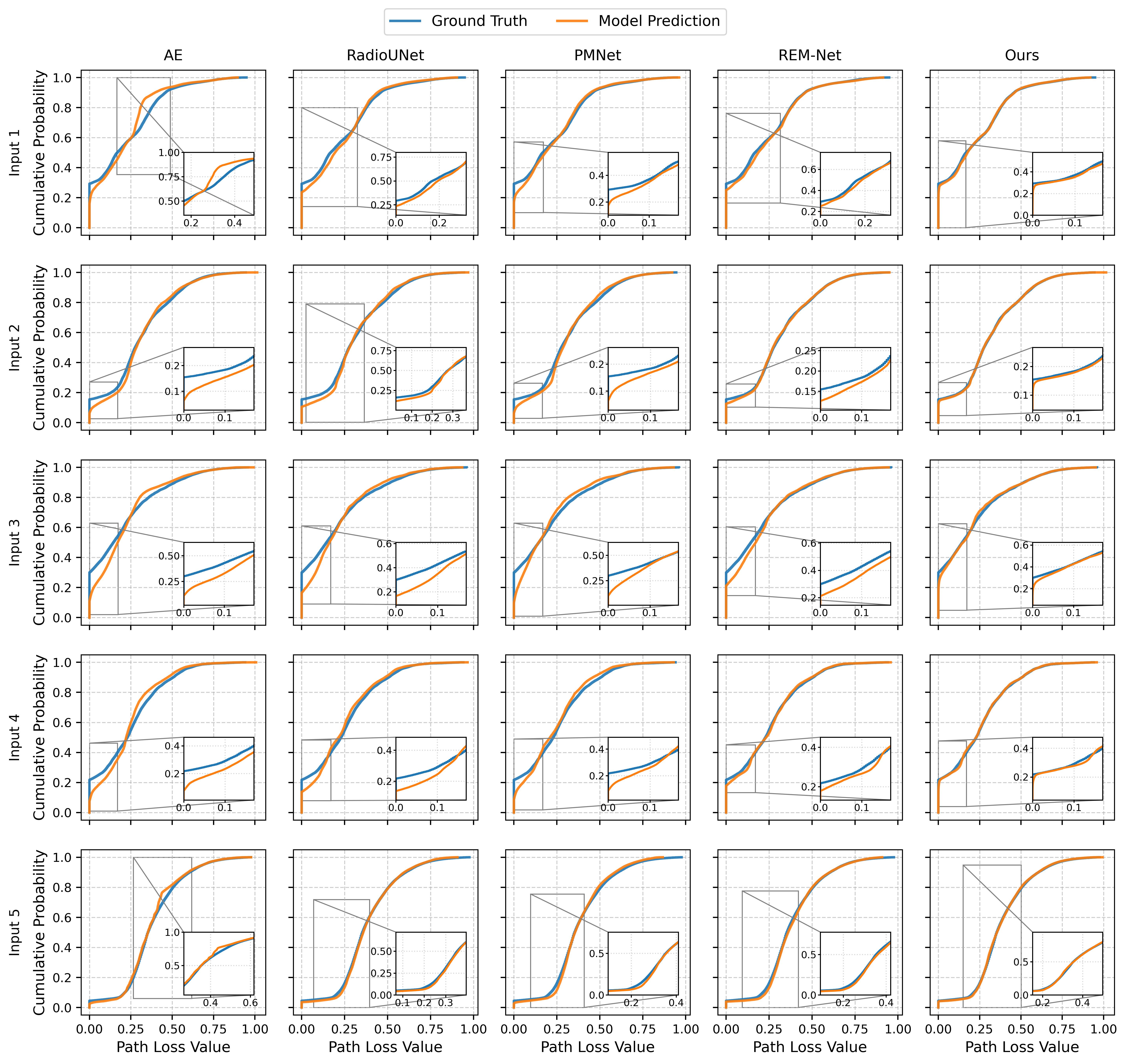}
    \caption{\added{Cumulative distribution of predicted path loss probabilities. PathFinder shows the closest alignment with Ground Truth, particularly in \textit{high-frequency} regions (0.0–0.25 range) where baseline models often deviate.}}
    \label{fig:comp_trail3}
\end{figure}

\added{To numerically substantiate this observation, we plotted the cumulative distribution of predicted path loss probabilities for all models across these samples, as shown in Fig. \ref{fig:comp_trail3}. To highlight the critical differences that are often obscured in standard plots, we enhanced Fig. \ref{fig:comp_trail3} with zoomed-in insets focusing on the high-frequency regions.}

\added{The AE model consistently exhibits significant discrepancies across most path loss intervals, matching the true labels only in environments with simpler building distributions, such as Input 5. In such sparse settings, signal propagation is dominated by line-of-sight components, making the task relatively straightforward for all methods and resulting in visually similar coverage curves.}

\added{However, in more complex environments with higher building density (\textit{e.g.}, Inputs 2, 3, and 4), where diffraction and reflection are prominent, the limitations of baseline models become evident. Specifically, RadioUNet, PMNet, and REM-Net struggle to achieve accurate predictions in the 0.0 to 0.25 path loss range. While the curves may appear visually close at a global scale, even minor deviations in this high-frequency range lead to substantial perceptual differences. In practice, these subtle numerical misalignments manifest as blurred boundaries and imprecise path loss estimation in densely built areas. In contrast, \mdl consistently maintains a distribution closer to the ground truth in these challenging high-frequency regions, resulting in a marked improvement in both numerical precision and perceptual boundary definition.}
\begin{table}[htbp]
  \centering
  \caption{{Comparison of different models' performance DS-RPP. \textbf{Bold} indicates the best model, while \uline{underline} indicates the second-best model.}}
  \resizebox{0.65\linewidth}{!}{
\begin{tabular}{c|cccccc}
\toprule
Model & MSE   & RMSE  & NMSE  & MSE-R & RMSE-R & NMSE-R \\
\midrule
AE    & 0.00478 & 0.069131 & 0.043575 & 0.005159 & 0.071818 & 0.038498 \\
RadioUNet & 0.003252 & 0.057026 & 0.029639 & 0.003313 & 0.057552 & 0.024713 \\
PMNet & 0.002399 & 0.048977 & 0.021872 & 0.002504 & 0.050035 & 0.018685 \\
REM-Net & \uline{0.002272} & \uline{0.047644} & \uline{0.020729} & \uline{0.002305} & \uline{0.047994} & \uline{0.017217} \\
PathFinder & \textbf{0.001096} & \textbf{0.033069} & \textbf{0.010004} & \textbf{0.001161} & \textbf{0.03404} & \textbf{0.008673} \\
\bottomrule
\end{tabular}}
  \label{tab:comparison}%
\end{table}%
\subsection{Comparison on Unseen Rural Dataset}
\added{As illustrated in Table \ref{tab:rural}, PathFinder consistently demonstrates superior performance across all six evaluation metrics. Notably, it achieves the lowest MSE ($0.1068$) and RMSE ($0.3263$), surpassing the strongest baseline, REM-Net (MSE: $0.1148$). Interestingly, while advanced models such as PMNet and RadioUNet exhibit competitive results on the DS-RPP task, their performance undergoes significant degradation on the Rural dataset, with PMNet's MSE escalating to $0.2596$. This discrepancy suggests that PathFinder is less susceptible to overfitting on specific architectural layouts and possesses stronger robust generalization.}

\begin{figure*}
    \centering
    \includegraphics[width=.9\linewidth]{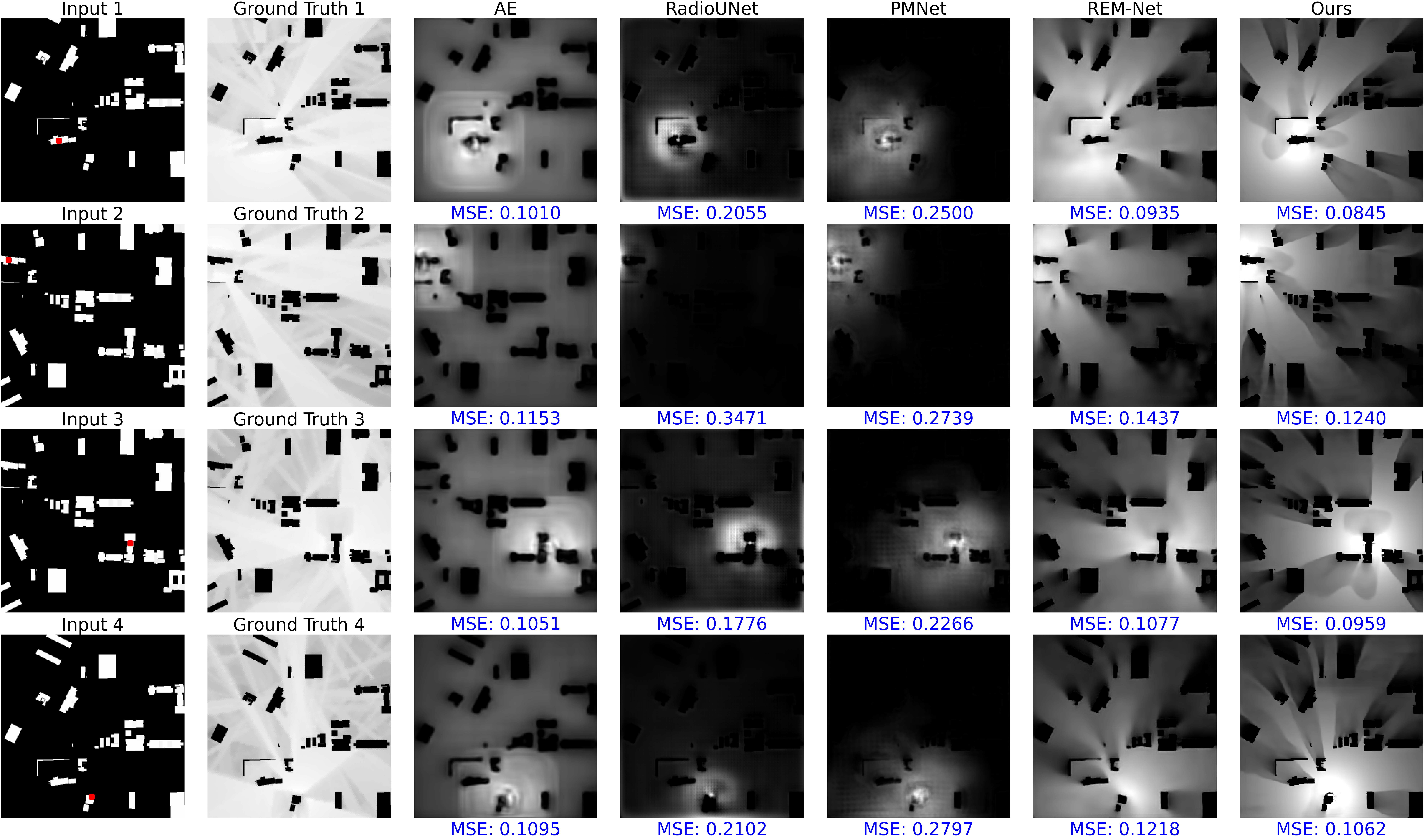}
    \caption{\added{Visual comparison of different models on the unseen Rural dataset. From left to right: input maps (red dots indicate Tx), Ground Truth, and predictions from various models.}}
    \label{fig:rural}
\end{figure*}

\added{However, given the divergent simulation settings of the Rural dataset, numerical results may fail to reflect the actual generalizability of the models, as seen in the superficial performance of the AE baseline. For instance, quantitative metrics alone might suggest that AE achieves performance comparable to that of REM-Net and PathFinder. To provide a more granular assessment, we conducted visualization experiments on four representative samples. As illustrated in Fig. \ref{fig:rural}, RadioUNet and PMNet struggle to generate accurate predictions, exhibiting the most pronounced degradation in generalization. While AE appears competitive numerically, the visualization reveals that it merely produces simplistic, circular path-loss boundaries centered at the transmitter, failing to capture complex propagation characteristics in distant regions. In contrast, REM-Net and PathFinder demonstrate superior robustness. Nevertheless, REM-Net still suffers from blurred and imprecise boundary predictions, particularly underperforming within intricate inter-building spaces.}
\begin{table}[htbp]
  \centering
  \caption{Comparison of model performance on the unseen Rural dataset. \textbf{Bold} indicates the best performance, while \uline{underline} indicates the second-best.}
   \resizebox{0.65\linewidth}{!}{
    \begin{tabular}{c|cccccc}
    \toprule
    Model & \multicolumn{1}{c}{MSE} & \multicolumn{1}{c}{RMSE} & \multicolumn{1}{c}{NMSE} & \multicolumn{1}{c}{MSE-R} & \multicolumn{1}{c}{RMSE-R} & \multicolumn{1}{c}{NMSE-R} \\
    AE    & \uline{0.110861 } & \uline{0.332849 } & \uline{0.309127 } & \uline{0.124505 } & \uline{0.352731 } & \uline{0.306588 } \\
    RadioUNet & 0.216644  & 0.465136  & 0.604718  & 0.240105  & 0.489693  & 0.591663  \\
    PMNet & 0.259646  & 0.509405  & 0.724408  & 0.285078  & 0.533768  & 0.702353  \\
    REM-Net & 0.114802  & 0.338248  & 0.320633  & 0.129394  & 0.359046  & 0.319004  \\
    PathFinder & \textbf{0.106828 } & \textbf{0.326344 } & \textbf{0.298455 } & \textbf{0.120286 } & \textbf{0.346247 } & \textbf{0.296670 } \\
    \bottomrule
    \end{tabular}}
  \label{tab:rural}%
\end{table}%
\subsection{Coverage Analysis}
In practical scenarios, models often only need to accurately predict critical areas. Regions with low signal strength can be considered noise \citep{RadioUNet}. Inspired by previous work \citep{RadioUNet}, we analyze model performance differences by visualizing a subset of samples. We consider the most important 40\%, 30\%, 20\%, and 5\% of path loss, respectively.

Table \ref{tab:comp_per} presents all models' performance regarding critical path loss area coverage. The final column shows our model's relative improvement over the second best model. As the importance factor increases (from 40\% to 5\%), \mdl's advantages become more pronounced. For the 40\% most important coverage area, performance improves by 34.46\% over REM-Net. For only the 5\% most important coverage area, \mdl outperforms REM-Net by 52.38\%. This demonstrates our model significantly enhances path loss prediction accuracy across different importance ratios.

Practically, predictions typically focus on the 20\% most important areas. Therefore, we visualize and analyze this scenario, as shown in Fig. \ref{fig:coverage80}. Visualizations for other importance levels appear in the appendix.

Inputs 1, 2, and 4 show that only \mdl predicts clear boundaries in areas farther from the transmitter. Other models, like AE and RadioUNet, exhibit substantial boundary prediction deficiencies. In complex building distribution areas (Input 3), only REM-Net and \mdl~accurately predict path loss. Other models fail to estimate path loss in these complex regions. For Input 5, models struggle with path loss predictions on the building side away from the transmitter. This is due to complex physical principles of signal propagation. REM-Net performs best among baselines, but its high frequency component predictions still fall short of \mdl. \mdl~shows a 51.87\% improvement, resulting in predictions perceptually closer to ground truth.
\begin{table}[htbp]
  \centering
  \caption{RMSE comparison of different models when addressing coverage areas of varying importance.}
  \resizebox{0.65\linewidth}{!}{
    \begin{tabular}{c|ccccc|c}
    \toprule
    Coverage/Model & AE    & RadioUNet & PMNet & REM-Net & Ours  & \multicolumn{1}{c}{Improvement} \\
    \midrule
    40    & 0.1249  & 0.0849  & 0.0756  & \uline{0.0582 } & \textbf{0.0382 } & 34.46\% \\
    30    & 0.1107  & 0.0717  & 0.0639  & \uline{0.0477 } & \textbf{0.0292 } & 38.77\% \\
    20    & 0.0721  & 0.0545  & 0.0488  & \uline{0.0350 } & \textbf{0.0197 } & 43.72\% \\
    10    & 0.0299  & 0.0306  & 0.0280  & \uline{0.0188 } & \textbf{0.0098 } & 48.09\% \\
    5    & 0.0142  & 0.0158  & 0.0150  & \uline{0.0096 } & \textbf{0.0046 } & 52.38\% \\
    \bottomrule
    \end{tabular}
    }
  \label{tab:comp_per}%
\end{table}%
\begin{figure*}
    \centering
    \includegraphics[width=0.95\linewidth]{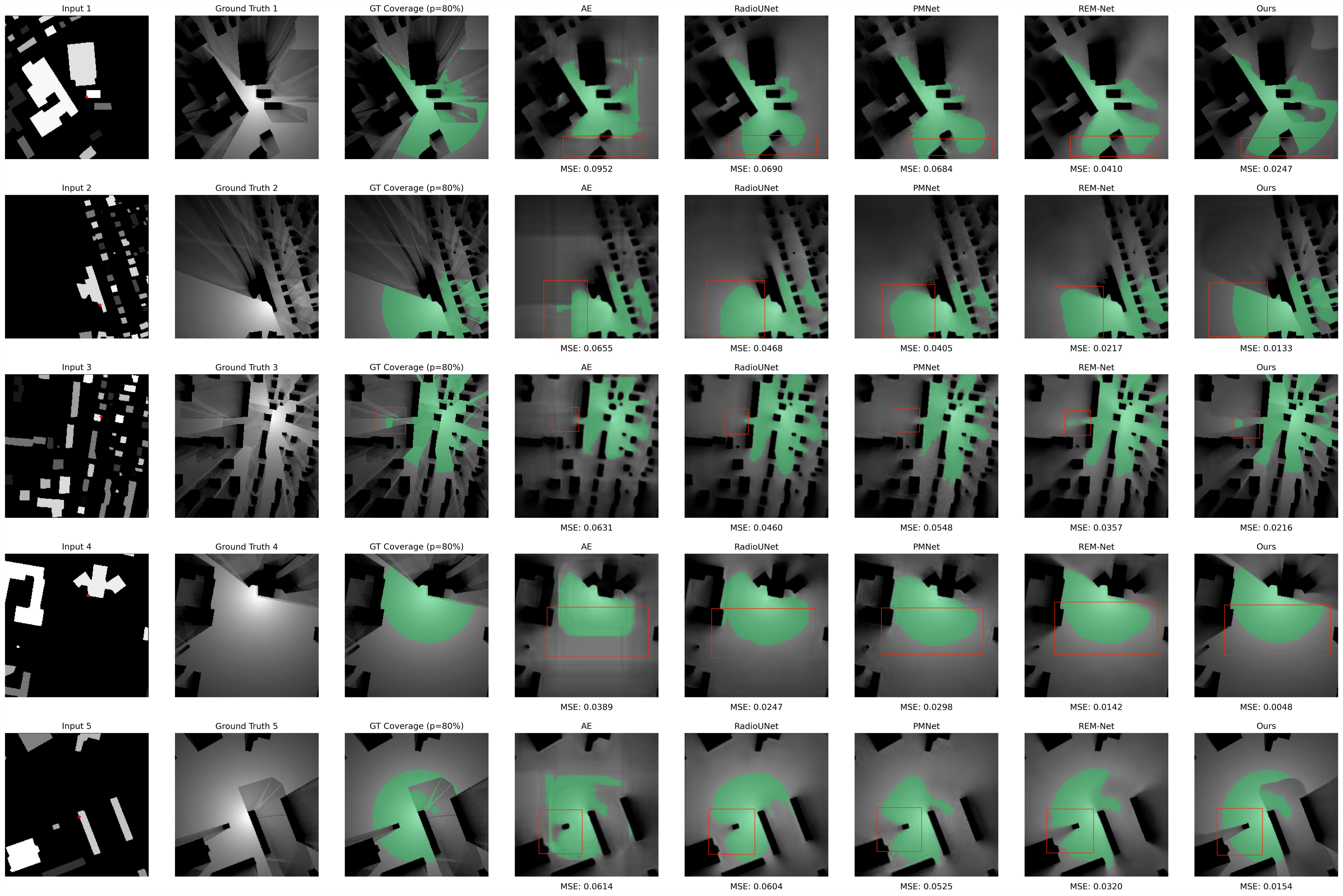}
    \caption{\added{Visualization of the top 20\% most important coverage areas. PathFinder demonstrates significantly clearer boundary predictions and higher fidelity in high-frequency components compared to state-of-the-art baselines.} }
    \label{fig:coverage80}
\end{figure*}

\subsection{Comparison on S2MT-RPP}
\begin{figure*}
    \centering
    \includegraphics[width=.9\linewidth]{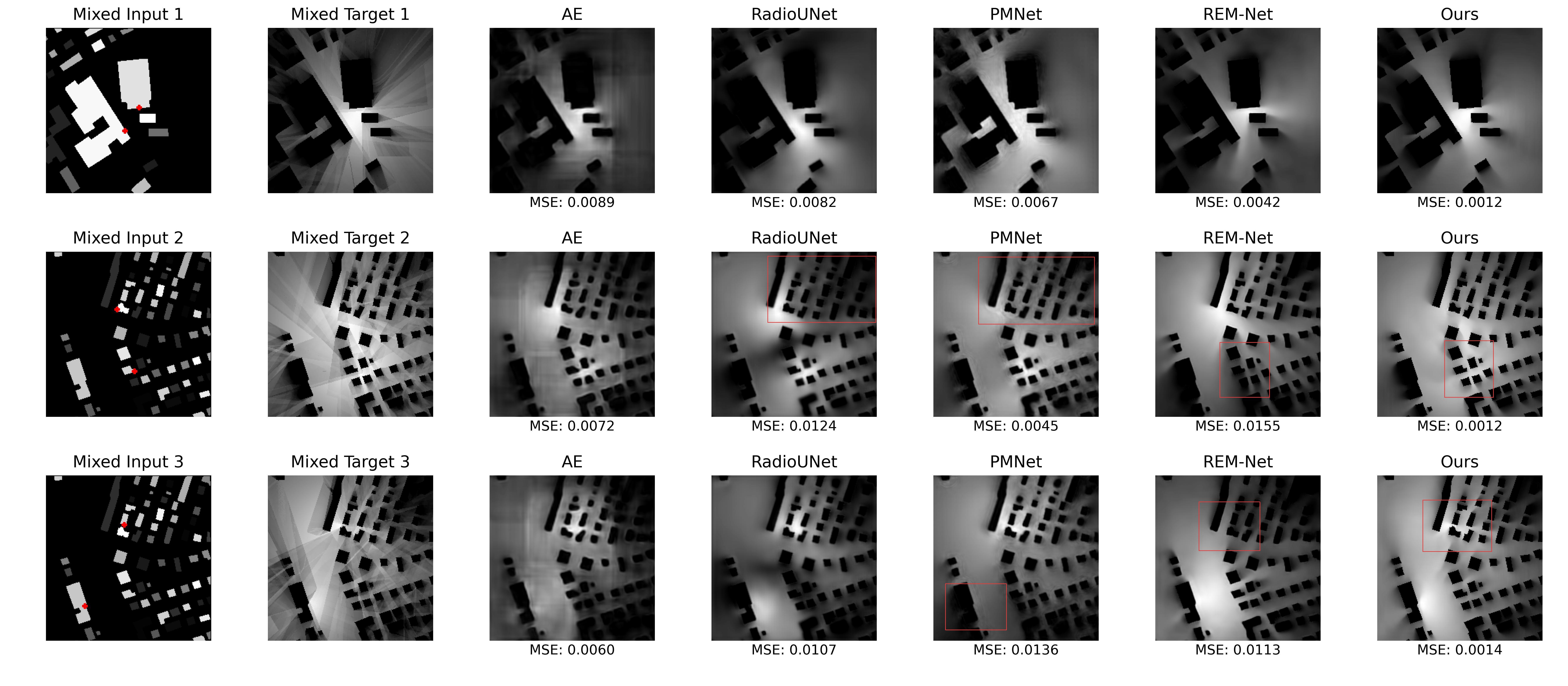}
    \caption{Visualization of performance of different models when the number of transmitters is 2.}
    \label{fig:mix2}
\end{figure*}

\begin{figure*}
    \centering
    \includegraphics[width=.9\linewidth]{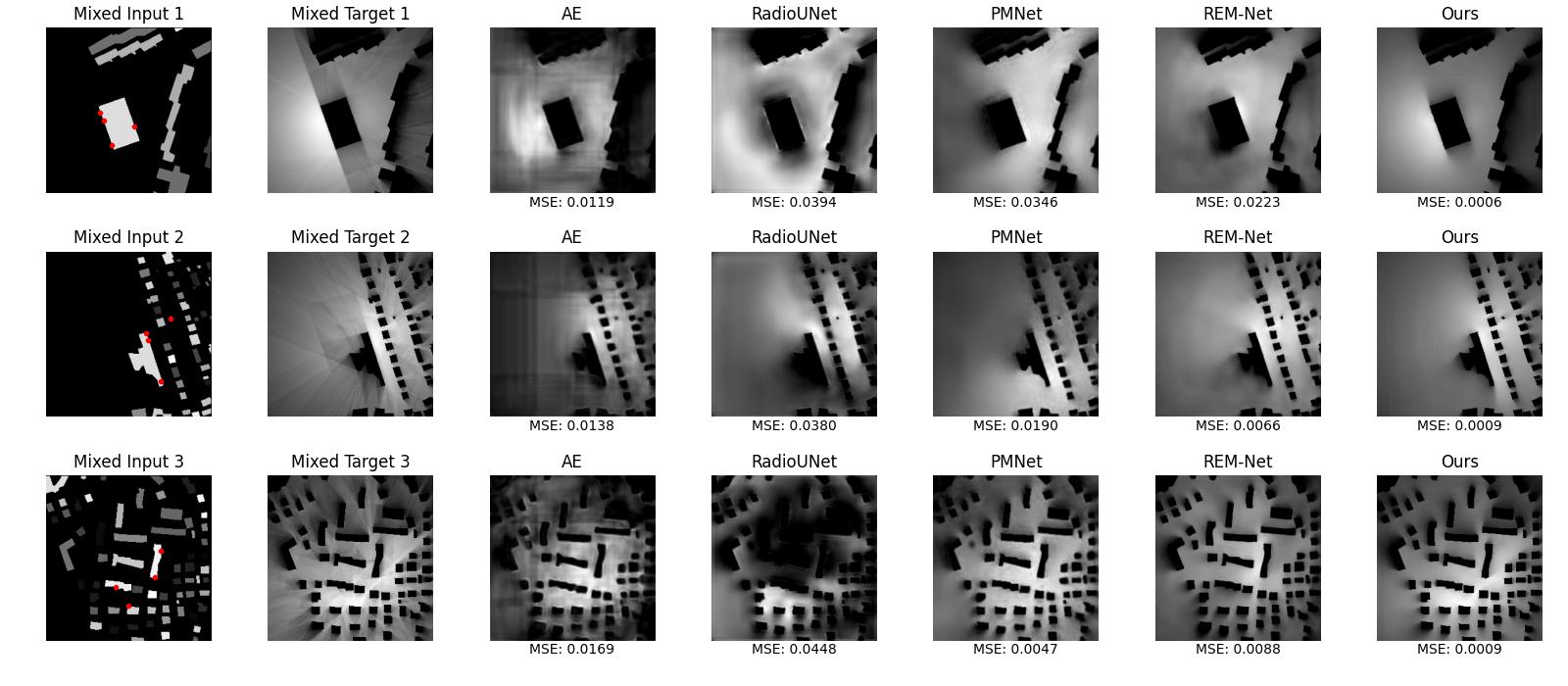}
    \caption{Visualization of performance of different models when the number of transmitters is 4.}
    \label{fig:mix4}
\end{figure*}

To analyze model performance with multiple transmitters, we first established a DS-RPP task with two transmitters by fixing the parameter \(\beta=0.5\) in Eq. (\ref{eq:mixup}) and omitting image enhancement. To further assess performance with varying transmitter counts, we modified Eq. (\ref{eq:mixup}) to average the effects of any number of \(\mS_i\) and \(\mY_i\) for S2MT-RPP. We evaluated models with 2 to 5 transmitters, presenting RMSE results in Table \ref{tab:mix_num} and visualizing cases with 2 and 4 transmitters in Figs. \ref{fig:mix2} and \ref{fig:mix4}, respectively.

Table \ref{tab:mix_num} shows that while baseline models like AE and RadioUNet experience increased loss and performance decline as transmitter count rises, more advanced models like PMNet and REM-Net maintain relatively consistent performance. Our model, however, demonstrates remarkable consistency across all transmitter counts, achieving a 59.35\% improvement over REM-Net even with five transmitters. This highlights the effectiveness of the proposed TOM strategy in enhancing our model's performance in distribution shift scenarios and its robust zero-shot capabilities.

Fig. \ref{fig:mix2} (2 transmitters) illustrates that when transmitters are close and unobstructed (Sample 1), most models (except AE) perform similarly. However, with complex building distributions, distant transmitters, or obstructions (Sample 2), RadioUNet and PMNet fail to accurately predict path loss in densely built areas and often only predict the effect of a single transmitter (Sample 3). REM-Net faces similar challenges, struggling with path loss predictions near transmitters in complex regions.

When the number of transmitters increases to four (Fig. \ref{fig:mix4}), the generalization capabilities of all baseline models, particularly AE and RadioUNet, largely fail. PMNet and REM-Net perform relatively better but still struggle to predict clear path loss boundaries even in straightforward building scenarios. In contrast, \mdl maintains a certain clarity in path loss boundaries across various situations, achieving a very low MSE of 0.0006 in simpler building distributions (Sample 1), indicating close alignment with true targets.

In summary, while current SOTA models show some ability to handle distribution shifts and exhibit zero-shot performance in multi-transmitter scenarios, their generalization in complex building distributions or with multiple transmitters is significantly inferior to that of \mdl.
\begin{table}[htbp]
  \centering
  \caption{Comparison of different models' performance under distribution shifts caused by varying numbers of transmitters.}
  \resizebox{0.65\linewidth}{!}{
    \begin{tabular}{c|ccccc|c}
    \toprule
    Tx-num/Model & AE    & RadioUNet & PMNet & REM-Net & Ours  & Improvement \\
    \midrule
    2     & 0.091157 & 0.097661 & \uline{0.090341} & 0.090515 & \textbf{0.03075} & 65.97\% \\
    3     & 0.114313 & 0.158482 & 0.092957 & \uline{0.083091} & \textbf{0.02873} & 65.43\% \\
    4     & 0.128223 & 0.176148 & 0.088597 & \uline{0.075713} & \textbf{0.02837} & 62.53\% \\
    5     & 0.134286 & 0.179481 & 0.082717 & \uline{0.070813} & \textbf{0.02879} & 59.35\% \\
    \bottomrule
    \end{tabular}}
  \label{tab:mix_num}%
\end{table}%
\subsection{Convergence Analysis of Different Models}\label{sec:convergence}
\begin{figure}
    \centering
    \resizebox{\linewidth}{!}{
    \includegraphics[width=0.9\linewidth]{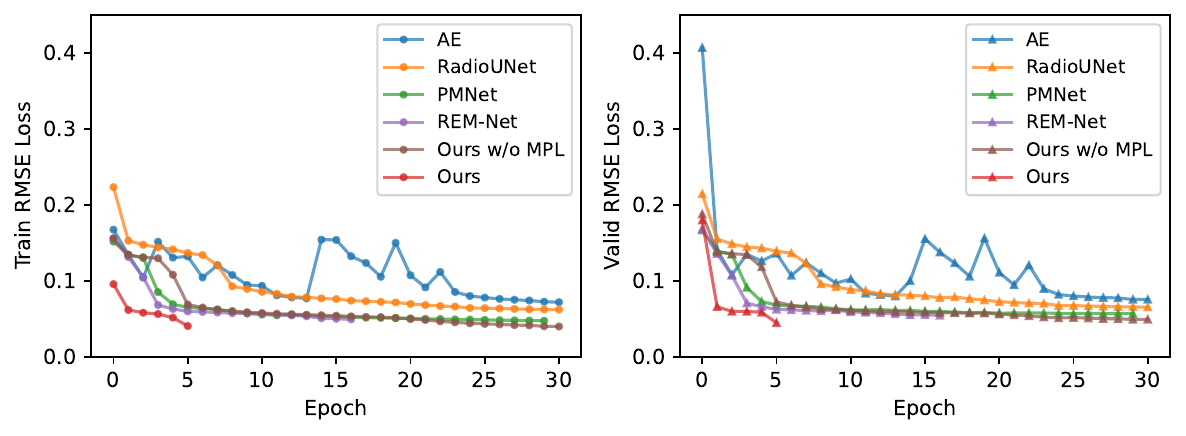}
    }
    \caption{Variation of RMSE loss during the training and validation phases across different models.}
    \label{fig:loss}
\end{figure}

To further investigate the reasons behind the differences in distribution shift handling capabilities across various models and to demonstrate that the proposed MPL optimization effectively aids model convergence, we analyze the convergence processes of different models. Fig. \ref{fig:loss} illustrates the changes in loss for all models during training and validation. "Ours w/o MPL" represents our model without MPL optimization, wherein MSE is used as a substitute loss for model optimization. In implementation, we set a maximum of 30 training epochs and employed an early stopping strategy, saving the best model when the validation loss no longer decreases. In our experiments, our model, along with PMNet and REM-Net, triggered this strategy.

From Fig. \ref{fig:loss}, it is evident that the AE model exhibits considerable fluctuations in loss during training, while the other models show minimal fluctuations. Additionally, REM-Net converges the fastest among all models, with a noticeable slowdown in loss reduction approximately after 3 epochs. Before utilizing MPL, the convergence speed of ``Ours w/o MPL'' is comparable to that of PMNet, with a similar slowdown in loss reduction after 5 epochs. After the implementation of MPL, our model demonstrates the fastest convergence rate among all models, achieving the lowest validation loss and quickly triggering the early stopping mechanism. This validates the effectiveness of MPL in significantly enhancing model convergence speed and enabling the model to reach lower loss values.

\subsection{Ablation Analysis}


\added{To evaluate the contribution of each component in PathFinder, we conduct ablation studies using the DS-RPP task metrics (Table \ref{tab:abl}).}

\added{Specifically, "Ours w/o MLA" removes the MLA module, while "Ours w/o TOM" uses only default augmentation. "Ours w/ MAE" and "Ours w/o MPL*" employ MAE or MSE as the sole training loss. To emphasize the MPL's contribution, these variants are trained for only 6 epochs at a $3 \times 10^{-4}$ learning rate. Notably, "Ours w/o MPL*" differs from "Ours w/o MPL" in Sec. \ref{sec:convergence}, as the latter requires 30 epochs and a lower learning rate ($3 \times 10^{-6}$) for stable convergence.}

\added{Results indicate that using MAE or MSE alone (Ours w/ MAE and Ours w/o MPL*) leads to slower convergence and instability; their performance at higher learning rates is significantly inferior to PathFinder. Removing the MLA module increases the NMSE-R from 0.008673 to 0.013328, underperforming even the "Ours w/o TOM" variant (0.012829). This confirms that explicitly guiding the model to focus on transmitter/receiver regions is vital for capturing environmental relationships. Furthermore, while "Ours w/o TOM" (NMSE 0.015128) outperforms "Ours w/o MLA", it remains behind PathFinder, demonstrating that signal-additivity-based augmentation substantially boosts generalization.}
\begin{table*}[htbp]
  \centering
  \caption{\added{Ablation study of PathFinder components on the DS-RPP task. The results demonstrate the impact of the Mask-Guided Low-rank Attention (MLA), Transmitter-Oriented Mixup (TOM), and Momentum Prediction Loss (MPL).}}
  \resizebox{0.65\linewidth}{!}{
    \begin{tabular}{c|cccccc}
    \toprule
    \multicolumn{1}{l|}{Model} & MSE   & RMSE  & NMSE  & MSE-R & RMSE-R & NMSE-R \\
    \midrule
    \multicolumn{1}{l|}{Ours w/o MLA} & \multicolumn{1}{r}{0.001777 } & \multicolumn{1}{r}{0.042131 } & \multicolumn{1}{r}{0.016224 } & \multicolumn{1}{r}{0.001784 } & \multicolumn{1}{r}{0.042211 } & \multicolumn{1}{r}{0.013328 } \\
    \multicolumn{1}{l|}{Ours w/o TOM} & \multicolumn{1}{r}{0.001657 } & \multicolumn{1}{r}{0.040685 } & \multicolumn{1}{r}{0.015128 } & \multicolumn{1}{r}{0.001717 } & \multicolumn{1}{r}{0.041416 } & \multicolumn{1}{r}{0.012829 } \\
    \multicolumn{1}{l|}{Ours w/ MAE} & \multicolumn{1}{r}{0.286838 } & \multicolumn{1}{r}{0.535535 } & \multicolumn{1}{r}{2.617612 } & \multicolumn{1}{r}{0.111683 } & \multicolumn{1}{r}{0.334141 } & \multicolumn{1}{r}{0.833947 } \\
    \multicolumn{1}{l|}{Ours w/o MPL*} & \multicolumn{1}{r}{0.286886 } & \multicolumn{1}{r}{0.535581 } & \multicolumn{1}{r}{2.618055 } & \multicolumn{1}{r}{0.111709 } & \multicolumn{1}{r}{0.334178 } & \multicolumn{1}{r}{0.834137 } \\
    \midrule
    PathFinder & 0.001096 & 0.033069 & 0.010004 & 0.0011605 & 0.034040 & 0.008673 \\
    \bottomrule
    \end{tabular}}
  \label{tab:abl}%
\end{table*}%

\subsection{Hyper-parameter Sensitivity of TOM}
\begin{figure}
    \centering
    \includegraphics[width=.7\linewidth]{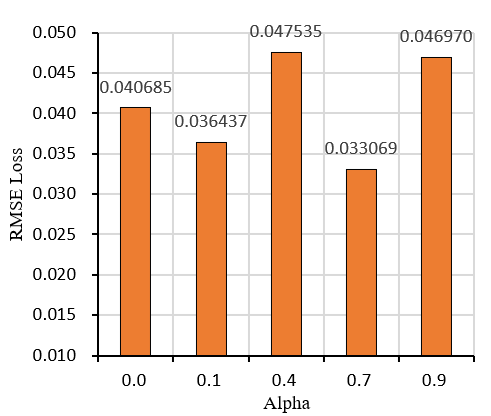}
    \caption{Hyper-parameter sensitivity analysis of the $\alpha$ parameter in the TOM strategy. Results are reported in RMSE on the DS-RPP testing set. $\alpha=0.7$ achieves the best balance between data diversity and physical realism.}
    \label{fig:alpha}
\end{figure}

\added{We analyzed the impact of the hyper-parameter $\alpha$ on the TOM strategy using the DS-RPP testing set. As shown in Fig.~\ref{fig:alpha}, the model performance follows a non-monotonic trend. While $\alpha=0.0$ represents the baseline without TOM augmentation, the optimal RMSE of 0.033069 is achieved at $\alpha=0.7$.}

\added{The performance degradation at $\alpha=0.4$ and $\alpha=0.9$ arises from distinct mechanisms. At $\alpha=0.4$, the mixing ratios are in a transition state that creates ambiguity. These samples are too far from original distributions to maintain basic propagation rules but lack the diversity required to learn the superposition principle. Consequently, the RMSE rises to 0.047535, which is worse than the baseline.}

\added{In contrast, $\alpha=0.9$ produces a nearly uniform distribution that frequently results in equal-intensity signal mixes. This causes signal saturation in complex environments where overlapping shadows and interference patterns become too intricate for the model to disentangle. Such over-mixed samples act as high-frequency label noise and lead to a performance drop to 0.046970.}

\added{Ultimately, $\alpha=0.7$ proves superior by generating dominant and subordinate signal combinations. This configuration effectively mimics real-world multi-transmitter interference and S2MT distribution shifts without introducing excessive complexity during training.}
\subsection{Complexity Analysis}
\label{sec:complexity}

\subsubsection{Complexity Analysis of MLA}
\added{Standard cross-attention mechanisms typically incur a computational complexity of $\mathcal{O}(N \cdot M \cdot d)$, where $N$ and $M$ are the sequence lengths of the target and source features, respectively, and $d$ is the feature dimension. In the context of RPP, if one attempts to compute the global correlation between the transmitter region and the receiver region directly (where $M \approx N = HW$), the complexity becomes quadratic $\mathcal{O}(N^2 d)$. This is computationally prohibitive for high-resolution path loss maps.}

\added{In contrast, our proposed Mask-Guided Low-Rank Attention (MLA) optimizes this process by introducing a low-rank bottleneck and a compact transmitter-aware prompt. By disentangling the transmitter prompt $\mathbf{P}$ from the environmental features, we utilize $\mathbf{P}$ as the intermediate bridge for attention computation. Let $n$ denote the number of transmitter prompts (where $n \ll N$) and $E$ be the low-rank embedding dimension. The computation involves linear projections of queries $\mathbf{Q}_{B}, \mathbf{Q}_{R} \in \mathbb{R}^{N \times E}$ and keys/values $\mathbf{K}_{P}, \mathbf{V}_{P} \in \mathbb{R}^{n \times E}$.}

\added{The cross-attention score calculation, e.g., $\mathbf{A}_{B} = \mathbf{Q}_{B} \mathbf{K}_{P}^{\top}$, involves the multiplication of an $(N \times E)$ matrix by an $(E \times n)$ matrix, resulting in a complexity of $\mathcal{O}(N n E)$. Similarly, the aggregation step $\mathbf{O}_1 = \text{Softmax}(\mathbf{A}_{B})\mathbf{V}_{P}$ incurs $\mathcal{O}(N n E)$. Since $n$ is a small constant and $E$ is a reduced dimension, the total complexity of our MLA is:}
\begin{equation}
\mathcal{C}_{MLA} = \mathcal{O}(N n E) \approx \mathcal{O}(N).
\end{equation}
\added{Thus, \mdl; reduces the global relationship modeling from $\mathcal{O}(N^2)$ to $\mathcal{O}(N)$ complexity with respect to the image resolution, ensuring superior scalability for large-scale urban environments.}

\subsubsection{Experimental Efficiency Analysis}
\added{Table \ref{tab:params} provides a comprehensive comparison of model parameters, computational cost (FLOPs), and inference speed (FPS).} 
\begin{table*}[htbp]
  \centering
  \caption{Comparison of model complexity and inference efficiency. FLOPs and FPS are measured using a $256 \times 256$ input resolution on an NVIDIA A800 GPU. }
    \begin{tabular}{c|ccc}
    \toprule
    Model & Params (M) & FLOPs (G) & FPS \\
    \midrule
    AE    & 19.66  & 108.82  & 316.63  \\
    RadioUNet & 13.27  & 19.13  & 276.14  \\
    PMNet & 33.34  & 82.79  & 99.99  \\
    REM-Net & 39.74  & 214.89  & 48.47  \\
    \midrule
    Ours  & 45.25  & 322.87  & 59.92  \\
    Ours w/o MLA & 44.08  & 303.77  & 96.73  \\
    Ours w/ std-attn & 46.96  & 348.75  & 4.00  \\
    \bottomrule
    \end{tabular}%
  \label{tab:params}%
\end{table*}%
\added{While PathFinder utilizes 45.25~M parameters to enhance its expressive capacity for complex environments, it maintains a high inference speed of 59.92~FPS. Notably, compared to the "Ours w/o MLA" variant, the integration of the MLA module adds only approximately \textbf{1~M} parameters, demonstrating its lightweight design. To further evaluate efficiency, we include the "Ours w/ std-attn" variant, which replaces the MLA with standard cross-attention to compute correlations between transmitter prompts and the entire receiver region.}

\added{Most significantly, PathFinder’s FPS is nearly \textbf{15$\times$} higher than the "Ours w/ std-attn" variant (59.92 vs. 4.00 FPS), empirically validating our theoretical $O(N)$ complexity analysis. 
Furthermore, although REM-Net seemingly presents lower theoretical complexity, its excessively wide architecture and reliance on memory-intensive parallel branches hinder effective hardware optimization and result in substantial memory overhead. Consequently, PathFinder surpasses the current SOTA REM-Net (48.47 FPS) in both inference speed and prediction accuracy. These results confirm that PathFinder achieves an optimal balance between modeling complexity and real-time execution.}

\section{Conclusion}
\added{This paper addresses 5G path loss estimation, extending to out-of-distribution prediction with multiple transmitters. Addressing prior research gaps, including insufficient environmental modeling, a single-transmitter focus, and neglected out-of-distribution path loss prediction, we introduce PathFinder. This model decouples building and transmitter features for active environmental modeling. It employs Mask-Guided Low-Rank Attention to learn global environmental features, independently focusing on the decoupled features.
Furthermore, we establish a novel S2MT-RPP task with the proposed Transmitter Oriented Mixup method. 
We empirically analyze various models' performance on this task and the reasons for suboptimal results. Experiments show that PathFinder achieves state-of-the-art performance across multiple distribution shift scenarios, showcasing robust generalization and zero-shot capabilities.}

\added{Despite these advancements, certain limitations remain. While PathFinder significantly outperforms baselines in accuracy, it utilizes a more number of parameters (45.25 M) compared to state-of-the-art models like PMNet or REM-Net.
Although our MLA module ensures a high inference speed of 59.92~FPS, the overall model size may still pose challenges for deployment on resource-constrained edge devices with limited memory. Future work aims to explore model compression techniques and more diverse real-world tasks, treating PathFinder as a foundational model to address practical challenges in evolving communication environments.}

\section*{Acknowledgment}
This work was supported in part by National Natural Science Foundation of China No. 92467109, U21A20478, National Key R\&D Program of China 2023YFA1011601, and the Major Key Project of PCL, China under Grant PCL2025A11 and PCL2025A13.




\bibliographystyle{elsarticle-num-names} 
\bibliography{main}

\newpage
\appendix

\section{Experiment}

\subsection{Evaluation Metric}\label{app:metric}
To evaluate the performance of different methods, we construct a total of six core metrics. The first three metrics are global coverage metrics, namely MSE, RMSE, and NMSE. The latter three metrics focus on the receiver area and are referred to as MSE-R, RMSE-R, and NMSE-R.

\textbf{Global Coverage Metrics}
1. Mean Squared Error (MSE) serves as a fundamental accuracy metric, directly reflecting the model's prediction error across the entire area. The calculation formula is given by:
\begin{equation}
    \text{MSE} = \frac{1}{HW} \sum_{i=1}^{HW} (\mY_i - \hat{\mY}_i)^2,
\end{equation}
where \(\mY_i\) and \(\hat{\mY}_i\) represent the true value and predicted value of the \(i\)-th pixel, respectively, and \(HW\) is the total number of pixels in the entire area. This metric amplifies the influence of outliers by taking the arithmetic mean of the squared errors, measuring the overall fitting error of the model across the entire study area.

2. Root Mean Squared Error (RMSE) normalizes the dimensions based on MSE, expressed as:
\begin{equation}
    \text{RMSE} = \sqrt{\frac{1}{HW} \sum_{i=1}^{HW} (\mY_i - \hat{\mY}_i)^2},
\end{equation}
This metric shares the same dimension as the true values, providing an intuitive reflection of the average deviation between the predicted and true values, making it more interpretable than MSE.

3. Normalized Mean Squared Error (NMSE) eliminates the dimensional influence through relative error representation, defined as:
\begin{equation}
    \text{NMSE} = \frac{\sum_{i=1}^{HW} (\mY_i - \hat{\mY}_i)^2}{\sum_{i=1}^{HW} \mY_i^2}.
\end{equation}
This metric transforms absolute errors into relative errors, making it more suitable for assessing scenarios with significant differences in path loss.

\textbf{Receiver Area-based Metrics}: The last three metrics focus on the receiver area \(\mathcal{M}' \in \{0,1\}^{HW \times 1}\) (where \(\mathcal{M}'_i=1\) indicates the \(i\)-th pixel belongs to the receiver area), enabling targeted evaluation of the specified region through a mask matrix. The specific forms correspond to the basic metrics but introduce area weights.

1. Mean Squared Error for Receiver (MSE-R):
\begin{equation}
    \text{MSE-R} = \frac{1}{|\mathcal{M}'|} \sum_{i=1}^{HW} \mathcal{M}'_i \cdot (\mY_i - \hat{\mY}_i)^2,
\end{equation}
where \(|\mathcal{M}'|\) denotes the total number of valid pixels in the receiver area (i.e., the number of pixels where \(\mathcal{M}'_i=1\)). This metric focuses on fitting errors in the receiver area through masked weighted averages, amplifying the influence of outliers in the target region.

2. Root Mean Squared Error for Receiver (RMSE-R):
\begin{equation}
    \text{RMSE-R} = \sqrt{\frac{1}{|\mathcal{M}'|} \sum_{i=1}^{HW} \mathcal{M}'_i \cdot (\mY_i - \hat{\mY}_i)^2} = \sqrt{\text{MSE-R}}.
\end{equation}
This metric normalizes dimensions based on MSE-R, aligning with the dimensions of the true values to intuitively reflect the average deviation between the predicted values and true values in the receiver area.

3. Normalized Mean Squared Error for Receiver (NMSE-R):
\begin{equation}
    \text{NMSE-R} = \frac{\sum_{i=1}^{HW} \mathcal{M}'_i \cdot (\mY_i - \hat{\mY}_i)^2}{\sum_{i=1}^{HW} \mathcal{M}'_i \cdot \mY_i^2}.
\end{equation}
This metric calculates relative errors in the receiver area, eliminating dimensional influences and normalizing regional differences, making it suitable for comparing model performance across different receiver scenarios.

In summary, the first category of global coverage metrics not only requires the model to focus on RPP estimation but also evaluates the model's overall fitting capability. In contrast, the second category of receiver area extended metrics solely considers the model's RPP ability, making it more task-specific.

\subsection{Comparison on DS-RPP}\label{app:Comparison on DS-RPP}

\begin{figure*}
    \centering
    \includegraphics[width=.95\linewidth]{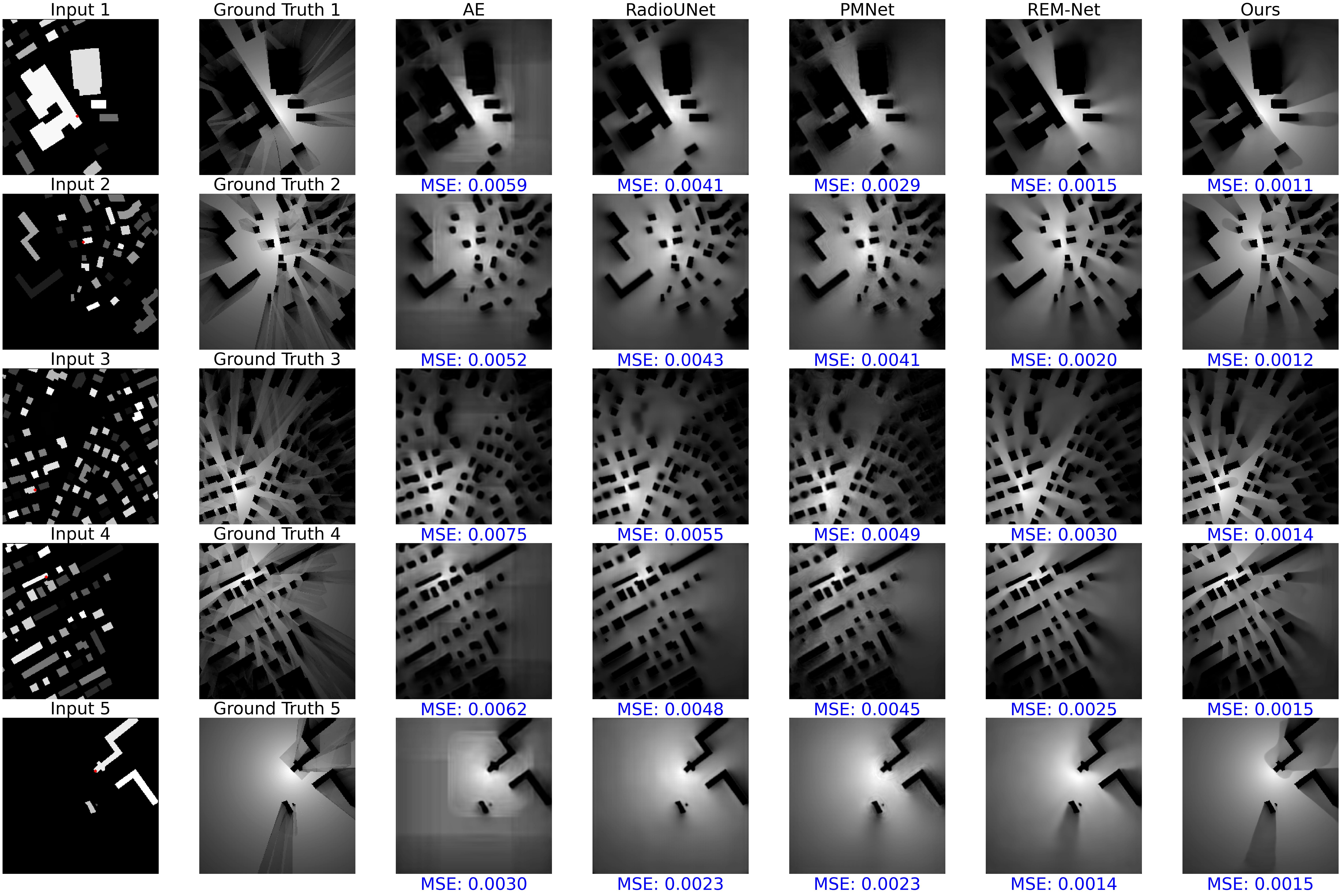}
    \caption{Visualization examples of different models in the DS-RPP task.}
    \label{app_fig:comp_trail1}
\end{figure*}

Fig. \ref{app_fig:comp_trail1} displays the prediction results of all baseline methods alongside our model across different samples.

\subsection{Coverage Analysis}\label{app:Coverage Analysis}

\begin{figure*}
    \centering
    \includegraphics[width=0.95\linewidth]{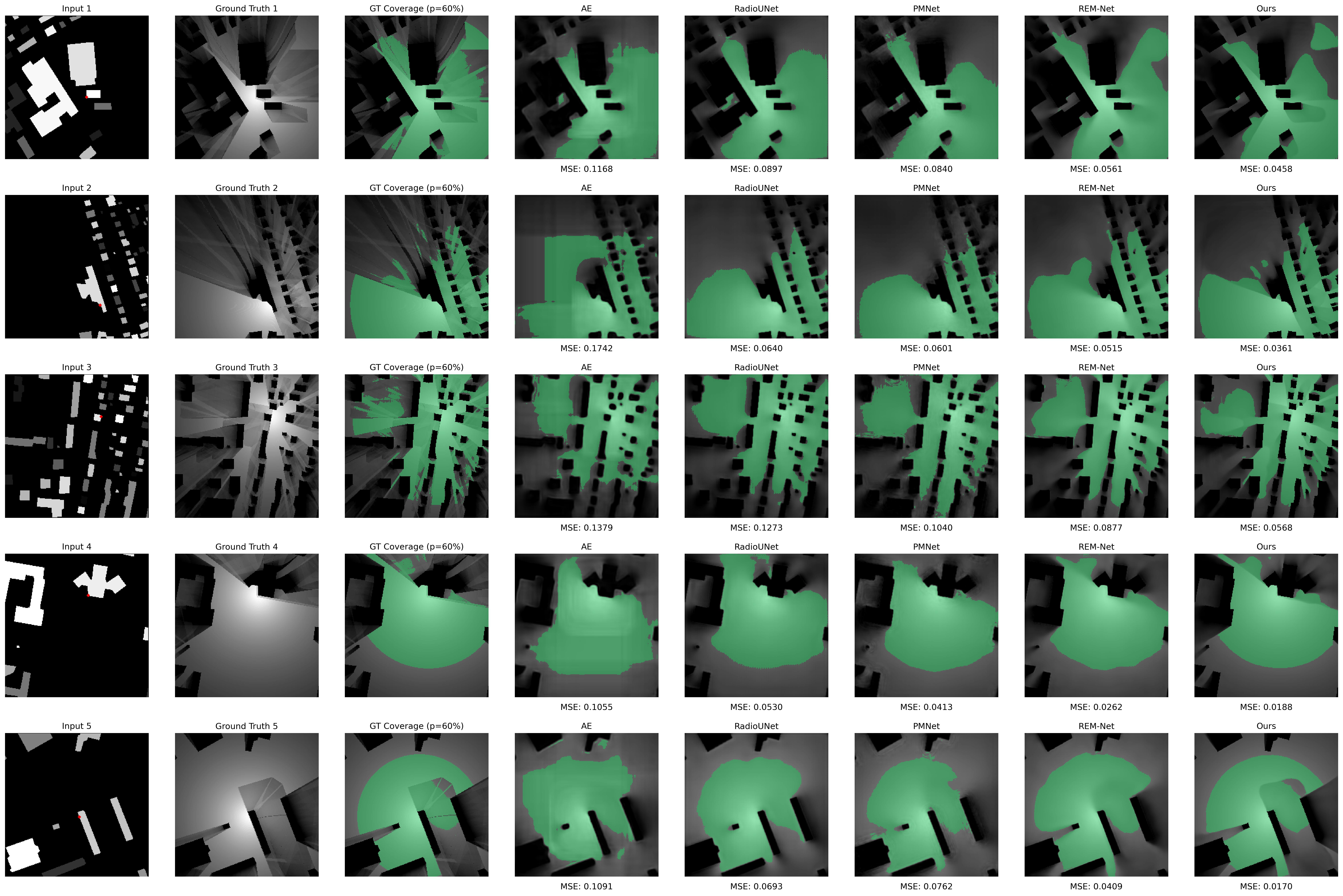}
    \caption{Visualization cases of different models when dealing with the most important 40\% of the coverage area. }
    \label{app_fig:coverage60}
\end{figure*}

\begin{figure*}
    \centering
    \includegraphics[width=0.95\linewidth]{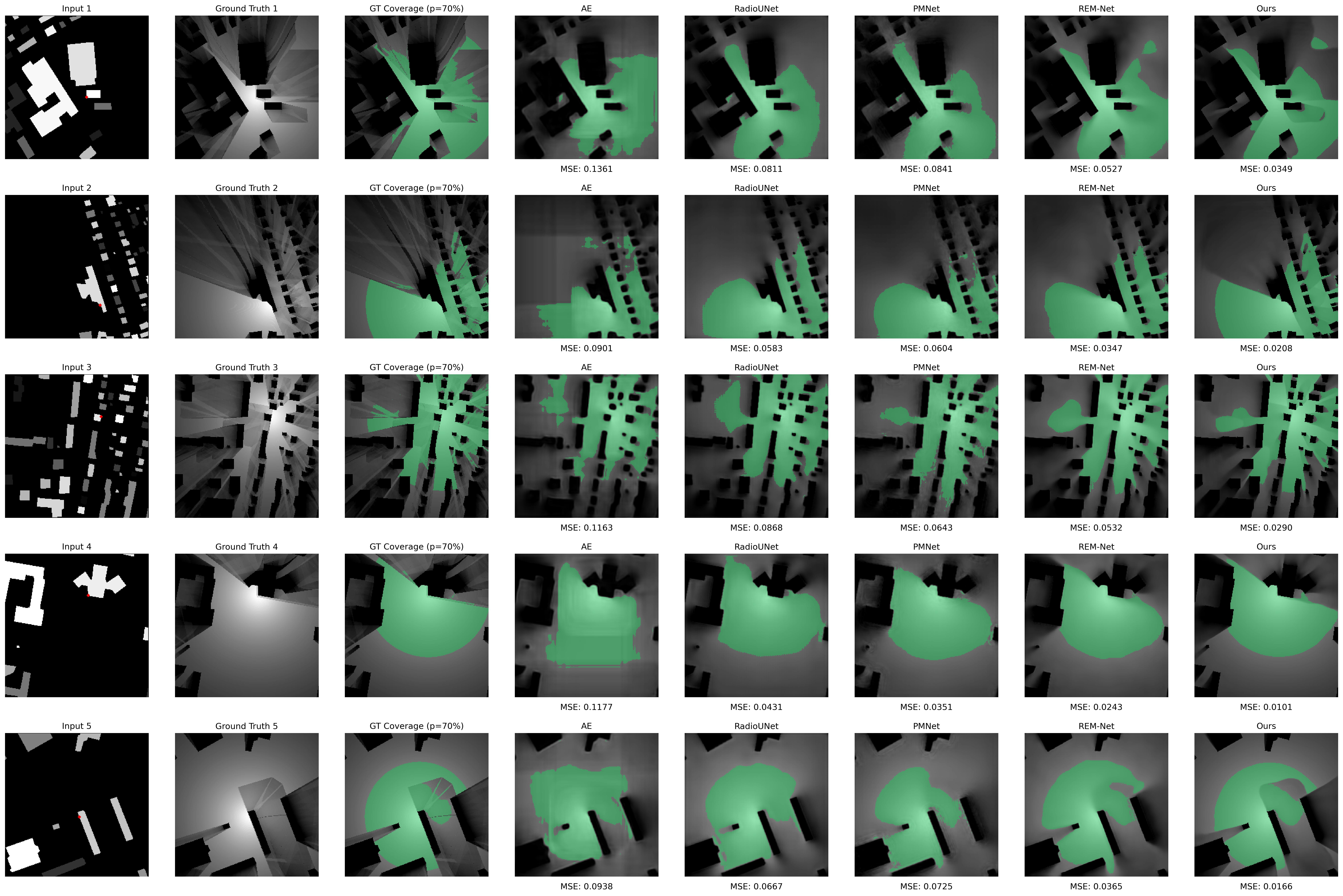}
    \caption{Visualization cases of different models when dealing with the most important 30\% of the coverage area. }
    \label{app_fig:coverage70}
\end{figure*}

\begin{figure*}
    \centering
    \includegraphics[width=0.95\linewidth]{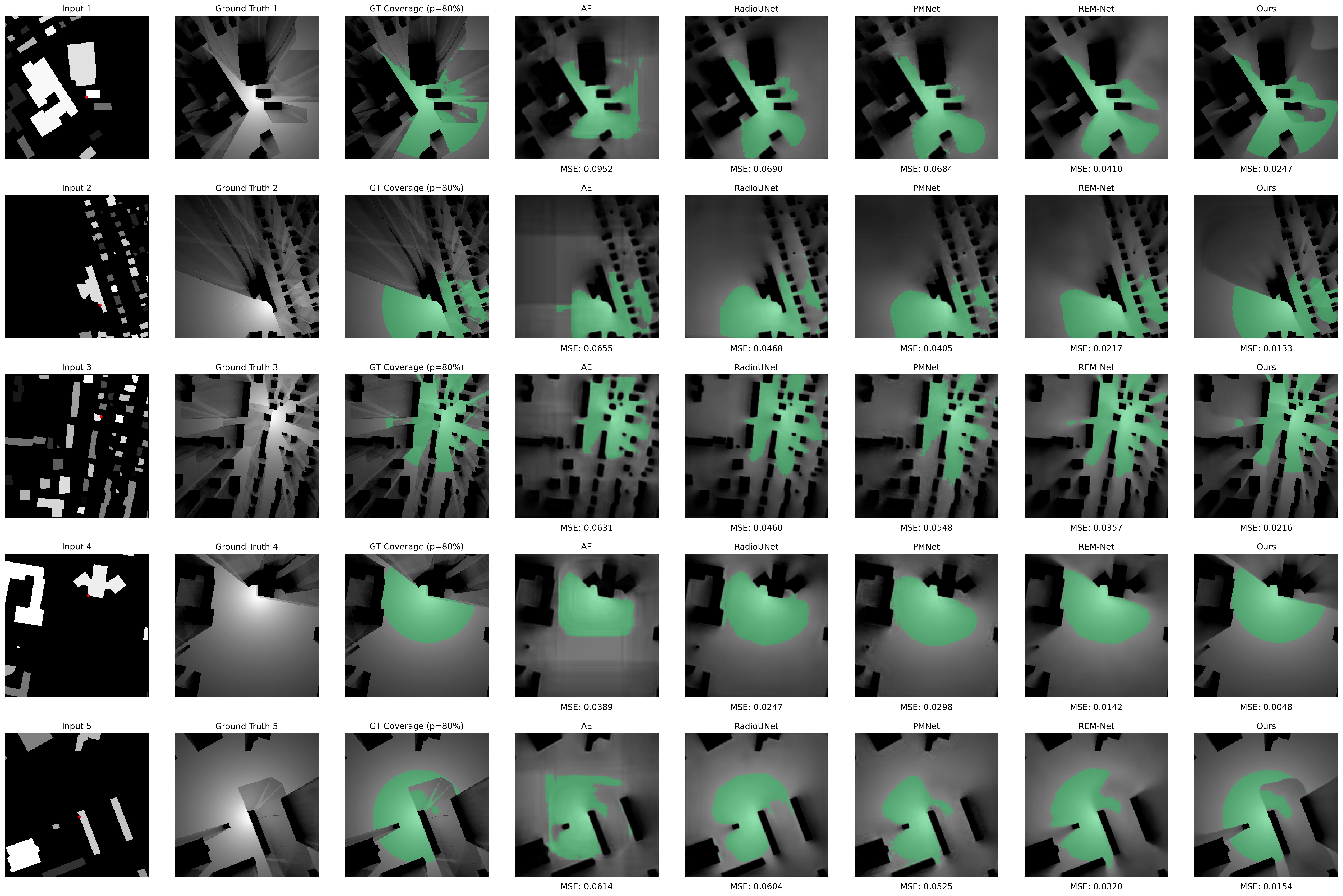}
    \caption{Visualization cases of different models when dealing with the most important 20\% of the coverage area. }
    \label{app_fig:coverage80}
\end{figure*}

\begin{figure*}
    \centering
    \includegraphics[width=0.95\linewidth]{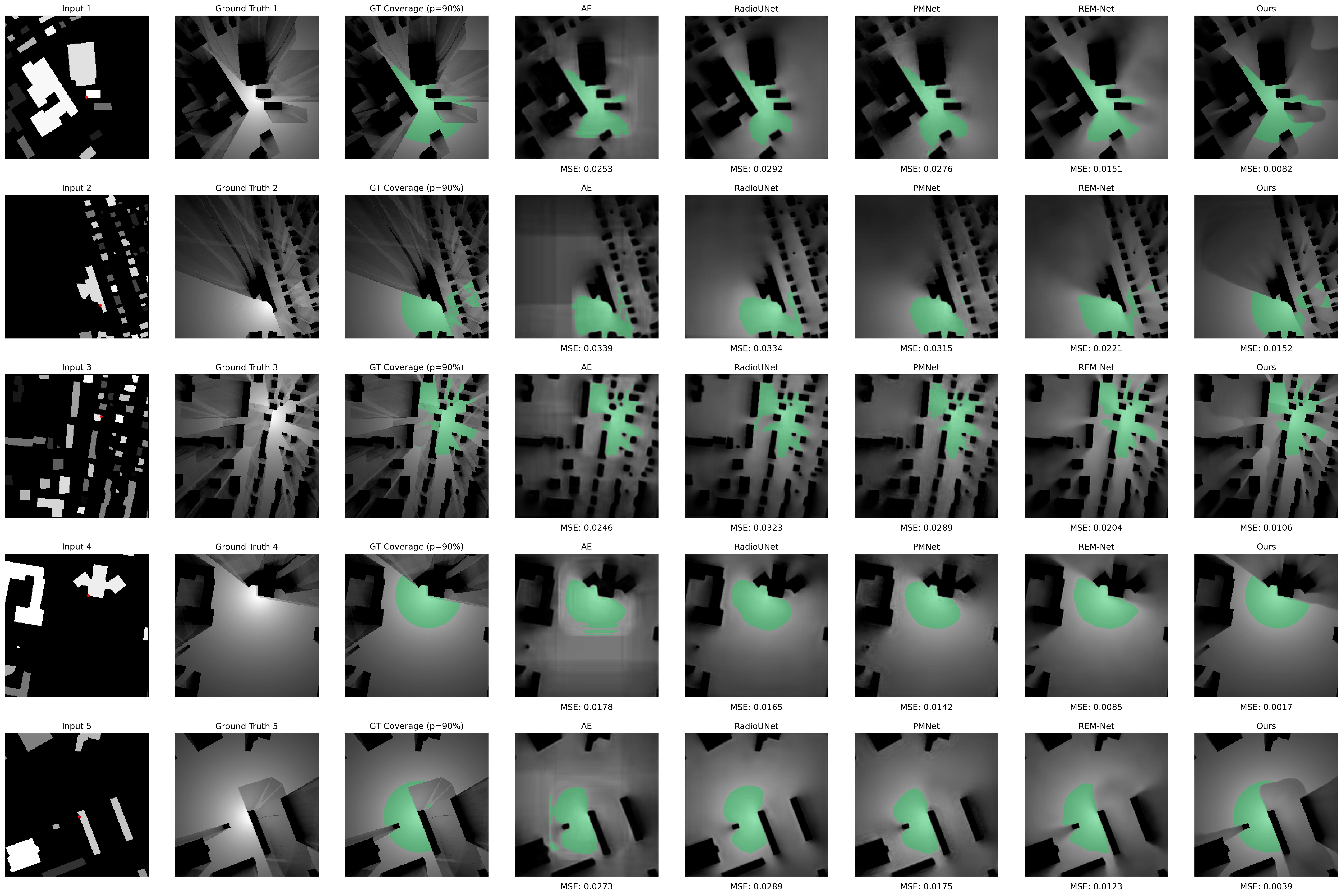}
    \caption{Visualization cases of different models when dealing with the most important 10\% of the coverage area. }
    \label{app_fig:coverage90}
\end{figure*}

\begin{figure*}
    \centering
    \includegraphics[width=0.95\linewidth]{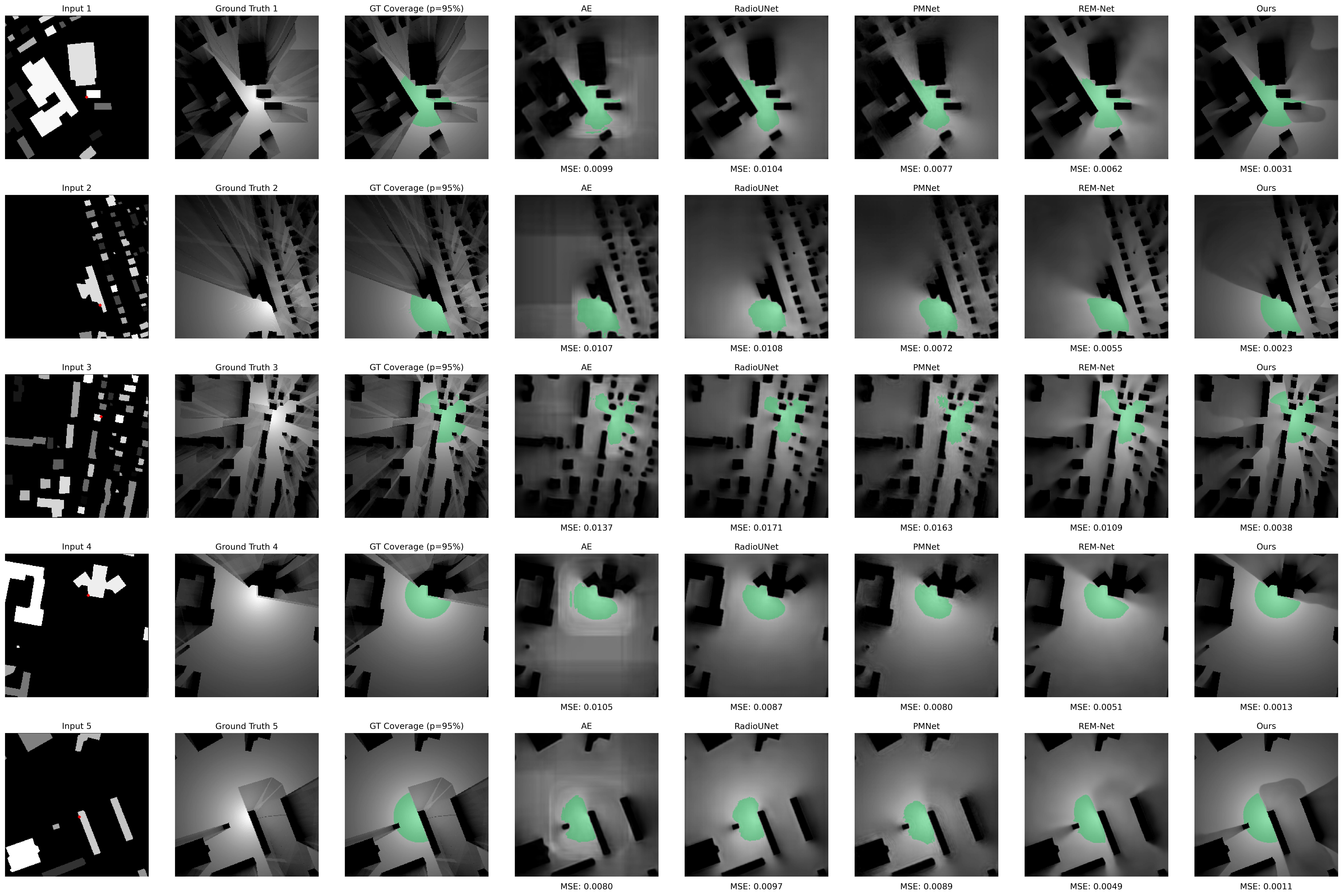}
    \caption{Visualization cases of different models when dealing with the most important 5\% of the coverage area. }
    \label{app_fig:coverage95}
\end{figure*}

Figs. \ref{app_fig:coverage60}, \ref{app_fig:coverage70}, \ref{app_fig:coverage80}, \ref{app_fig:coverage90} and \ref{app_fig:coverage95} visualize the results of different models when the importance of the coverage area is 40\%, 30\%, 20\%, 10\% and 5\%, respectively.

\begin{table}[htbp]
  \centering
  \caption{Comparison of different models' performance under distribution shifts caused by varying numbers of transmitters.}
  \resizebox{\linewidth}{!}{
    \begin{tabular}{c|c|ccccc|c}
    \toprule
    Tx-num/Model & Metric & \multicolumn{1}{c}{AE} & \multicolumn{1}{c}{RadioUNet} & \multicolumn{1}{c}{PMNet} & \multicolumn{1}{c}{REM-Net} & \multicolumn{1}{c|}{Ours} & Improvement \\
    \midrule
    \multirow{3}[2]{*}{2} & MSE   & 0.008311 & 0.009541 & \uline{0.008171} & 0.008204 & \textbf{0.00095} & 88.43\% \\
          & RMSE  & 0.091157 & 0.097661 & \uline{0.090341} & 0.090515 & \textbf{0.03075} & 65.97\% \\
          & NMSE  & 0.133632 & 0.15345 & \uline{0.131345} & 0.131917 & \textbf{0.0152} & 88.43\% \\
    \midrule
    \multirow{3}[2]{*}{3} & MSE   & 0.01307 & 0.025124 & 0.008654 & \uline{0.006909} & \textbf{0.000825} & 88.05\% \\
          & RMSE  & 0.114313 & 0.158482 & 0.092957 & \uline{0.083091} & \textbf{0.028727} & 65.43\% \\
          & NMSE  & 0.220913 & 0.424658 & 0.146268 & \uline{0.116794} & \textbf{0.013955} & 88.05\% \\
    \midrule
    \multirow{3}[2]{*}{4} & MSE   & 0.016445 & 0.031034 & 0.00786 & \uline{0.005737} & \textbf{0.000805} & 85.97\% \\
          & RMSE  & 0.128223 & 0.176148 & 0.088597 & \uline{0.075713} & \textbf{0.028369} & 62.53\% \\
          & NMSE  & 0.28556 & 0.538926 & 0.136471 & \uline{0.099624} & \textbf{0.013983} & 85.96\% \\
    \midrule
    \multirow{3}[2]{*}{5} & MSE   & 0.018037 & 0.032217 & 0.00685 & \uline{0.005018} & \textbf{0.00083} & 83.48\% \\
          & RMSE  & 0.134286 & 0.179481 & 0.082717 & \uline{0.070813} & \textbf{0.02879} & 59.35\% \\
          & NMSE  & 0.318152 & 0.568353 & 0.120816 & \uline{0.088533} & \textbf{0.01463} & 83.48\% \\
    \bottomrule
    \end{tabular}}
  \label{tab:comp_mix_app}%
\end{table}%

\end{document}

%% file: math_command.tex
\usepackage{amsmath,amsfonts,bm}

\def\eqref#1{equation~\ref{#1}}

\def\1{\bm{1}}

\def\rp{{\textnormal{p}}}

\def\rs{{\textnormal{s}}}

\def\mA{{\bm{A}}}
\def\mB{{\bm{B}}}

\def\mI{{\bm{I}}}

\def\mK{{\bm{K}}}

\def\mM{{\bm{M}}}

\def\mO{{\bm{O}}}
\def\mP{{\bm{P}}}
\def\mQ{{\bm{Q}}}
\def\mR{{\bm{R}}}
\def\mS{{\bm{S}}}

\def\mV{{\bm{V}}}

\def\mX{{\bm{X}}}
\def\mY{{\bm{Y}}}
\def\mZ{{\bm{Z}}}

\DeclareMathAlphabet{\mathsfit}{\encodingdefault}{\sfdefault}{m}{sl}
\SetMathAlphabet{\mathsfit}{bold}{\encodingdefault}{\sfdefault}{bx}{n}
\newcommand{\tens}[1]{\bm{\mathsfit{#1}}}

\def\tV{{\tens{V}}}
\def\tW{{\tens{W}}}

\def\sS{{\mathbb{S}}}

\def\sY{{\mathbb{Y}}}

